\RequirePackage{fix-cm}

\documentclass[twocolumn]{svjour3}        

\usepackage{graphicx}
\usepackage{multirow}
\usepackage{pifont}
\newcommand{\xmark}{\ding{55}}

\usepackage{color}
\usepackage{amsmath}
\usepackage{amssymb}
\usepackage{booktabs}

\begin{document}

\title{Multi-modal Prompting for Low-Shot Temporal Action Localization}

\titlerunning{Multi-modal Prompting for low-shot TAL} 

\author{Chen Ju \and
        Zeqian Li \and 
        Peisen Zhao \and 
        Ya Zhang \and    
        Xiaopeng Zhang \and  \\
        Qi Tian \and  
        Yanfeng Wang \and  
        Weidi Xie }

\authorrunning{Chen Ju et al.} 

\institute{{Chen Ju, Zeqian Li, Ya Zhang, Yanfeng Wang, Weidi Xie}  \at
              CMIC, Shanghai Jiao Tong University, Shanghai \\
              \email{\{ju\_chen, lzq0103, ya\_zhang\}@sjtu.edu.cn   \\ \{wangyanfeng, weidi\}@sjtu.edu.cn}           
           \and
           {Peisen Zhao, Xiaopeng Zhang, Qi Tian} \at
              Huawei Cloud \& AI, Shenzhen, Guangdong \\
             \email{\{pszhao93, zxphistory, tianqi1\}@gmail.com}
}

\date{Received: date / Accepted: date}

\maketitle

\begin{abstract}
\hspace{1pt} In this paper, we consider the problem of temporal action localization under low-shot~(zero-shot \& few-shot) scenario, with the goal of detecting and classifying the action instances from arbitrary categories within some untrimmed videos, even not seen at training time. We adopt a Transformer-based two-stage action localization architecture with class-agnostic action proposal, followed by open-vocabulary classification. 
We make the following contributions. 
\textbf{First}, to compensate image-text foundation models with temporal motions, we improve category-agnostic action proposal by explicitly aligning embeddings of optical flows, RGB and texts, which has largely been ignored in existing low-shot methods. 
\textbf{Second}, to improve open-vocabulary action classification, we construct classifiers with strong discriminative power, {\em i.e.}, avoid lexical ambiguities. To be specific, we propose to prompt the pre-trained CLIP text encoder either with detailed action descriptions~(acquired from large-scale language models), or visually-conditioned instance-specific prompt vectors.
\textbf{Third}, we conduct thorough experiments and ablation studies on THUMOS14 and ActivityNet1.3, demonstrating the superior performance of our proposed model, outperforming existing state-of-the-art approaches by one significant margin.

\keywords{Vision-Language Foundation Models \and Prompt Tuning \and Low-shot Video Understanding}
\end{abstract}

\begin{figure*}[t]
\begin{center}
\vspace{0.2cm}
\includegraphics[width=\textwidth] {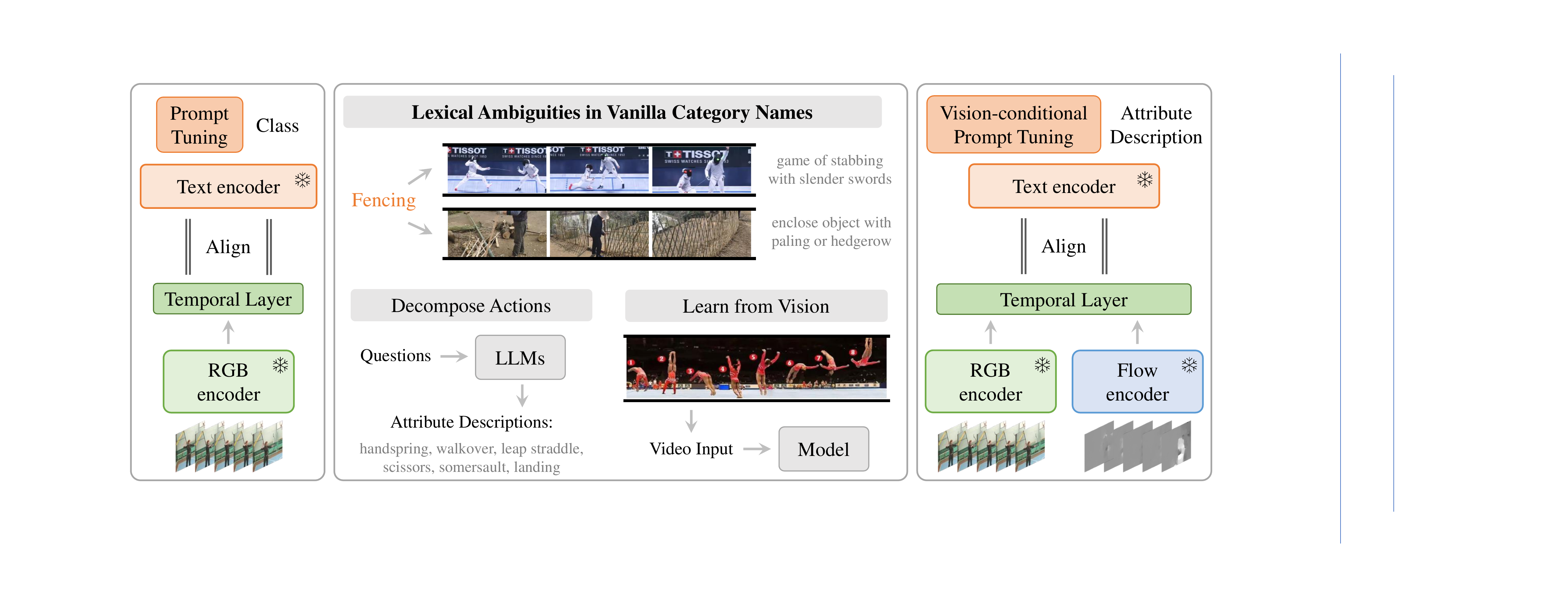}
\end{center}
\vspace{-0.2cm}
\caption{\textbf{Left}: Existing methods add prompt tuning and temporal layer to foundation models. \textbf{Middle}: The main challenge is lexical ambiguities in vanilla category names. To disambiguate text-based category names, we decompose actions by prompting large language models for various action attribute descriptions. For cases where it may be difficult to give comprehensive detail descriptions, we further propose vision-conditional prompting to learn from input videos. \textbf{Right}: Our overall framework.}
\vspace{0.2cm}
\label{fig:intro}
\end{figure*}

\section{Introduction}
\label{intro}
Temporal action localization~(TAL) aims to localise and classify action instances in untrimmed long videos and is an indispensable part of video understanding~\cite{shu2015joint,zhu2016learning,ke2010volumetric}. Despite great progress has been made by training deep neural networks on large-scale datasets~\cite{caba2015activitynet,zhao2019hacs}, existing research mainly considers a closed-set scenario, where action categories remain identical at training and inference stage. Such an assumption is clearly an oversimplification for deployment scenarios, hence limiting its practical uses.

In the recent literature, another line of research~\cite{ju2022prompting,nag2022zero} considers a more challenging problem, that requires the vision system to handle both seen and unseen categories, with low-shot (zero or only few) examples at inference time, this problem is often termed as open-vocabulary temporal action localization. To tackle the problem, existing studies~\cite{ju2022prompting,nag2022zero,luo2022clip4clip} take inspiration from large-scale foundational models~\cite{Radford21,Jia21,yao2021filip}, casting the problem of action classification in the form of cross-modal retrieval, {\em i.e.}, for one action videos, searching its closest category embedding in text form~({\em e.g.}, ``an action video of class''). However, such a design potentially suffers from the lexical ambiguities, as multiple actions may share category names, despite its differing visual appearance. For example in Figure~\ref{fig:intro}, ``fencing'' could either refer to ``the game of stabbing with slender steel swords in protective clothing'' or ``enclose or separate something using paling or hedgerow'', directly encoding the category names would therefore be unable to distinguish these two concepts.

To alleviate the above challenge, in this paper, we consider two ideas to enrich the discriminative power of action classifiers constructed from a pre-trained text encoder, via natural language descriptions, or visually conditioned prompting. To be specific, (1) to disambiguate the text-based category names, we construct an automatic pipeline to source detailed attribute descriptions for action categories by prompting large-scale language models~(LLMs)~\cite{Brown20,zhang2022opt}. Along this line of idea, we propose three types of attribute question templates, to encode the salient objects, event fields, and interactions for the query actions. For instance, while prompting LLMs using the template: ``What tools are needed for [skiing]?'', it outputs ``ski board, ski stick, ski goggles, helmet'', hence giving additional cues to enhance the discrimination of the subsequently generated action classifier; 
(2) in cases that could be difficult to provide comprehensive detail descriptions, we further propose vision-conditional prompting, that enables to extract action details from RGB and Flow embeddings, as instance-specific prompt vectors for the pre-trained text encoder. As an example demonstrated in Figure~\ref{fig:intro}, to construct the action classifier for ``competitive artistic gymnastics'', we can either encode the detailed language descriptions~(handspring, walkover, leap straddle, scissors, somersault, etc.), or simply calculate the visual embeddings of the given video, and prompt the text encoder along with action categories.

Inheriting empirical observations that optical flows can generally bring impressive category-agnostic detection (action proposals), {\em i.e.}, large flow normally indicates the existence of actions, we also explicitly incorporate optical flows into the visual representations~\cite{carreira2017quo}, and then align RGB, Flow, and text modalities into one shared embedding space. When evaluating on two standard benchmarks: THUMOS14 and ActivityNet1.3, our model significantly outperforms existing competitive methods, sometimes by over 10\% across few-shot and zero-shot scenarios. We also conduct thorough ablation studies to reveal the effectiveness of each component, both quantitatively and qualitatively.

\section{Related Work} \label{sec:related work}
\vspace{-0.2cm}
{\noindent \bf Vision-Language Pre-training.} In the early literature, \cite{Mori99,Frome13,Weston11} explore jointly training image-text embeddings using paired text documents. Recently, some studies have further scaled up the training with large-scale web data to form ``the \textbf{foundation} models'', {\em e.g.}, CLIP~\cite{Radford21}, ALIGN~\cite{Jia21}, Florence~\cite{yuan2021florence}, FILIP~\cite{yao2021filip}, VideoCLIP~\cite{xu2021videoclip}, and LiT~\cite{zhai2022lit}. These foundation models usually contain one visual encoder and one textual encoder, which are trained using simple noise contrastive learning for powerful cross-modal representations. They have shown promising potential in many tasks, such as image classification and detection, action recognition, and retrieval. In this paper, we use CLIP for low-shot temporal action localization, but the same technique should be applicable to other foundation models as well.

\vspace{0.1cm}
{\noindent \bf Prompting} refers to leveraging input instructions to steer foundation models for desired outputs. In the NLP domain, early papers~\cite{Gao21,Jiang20,Timo21,Shin20} focus on handcrafted prompt templates. To avoid labor and increase flexibility, some studies~\cite{Lester21,li21-prefixtuning,li2021prefix} propose learnable prompt tuning at the textual stream, showing strong low-shot generalization. In the CV domain, some recent papers~\cite{zhou2019learn,zhou2022conditional,ju2022prompting} introduce such randomly initialized prompt tuning to handle visual tasks, {\em e.g.}, image understanding~\cite{zhu2022prompt,lu2022prompt,yang2022learning,ma2023diffusionseg} and video understanding~\cite{jia2022visual,nag2022zero,ni2022expanding}. However, these studies ignore lexical ambiguity of category names, and cases that are not easy to describe in text. This paper designs novel conditional prompt tuning and language descriptions from LLMs, to solve these issues.

\vspace{0.1cm}
{\noindent \bf Closed-set Temporal Action Localization} considers to detect and classify action instances from one pre-defined category list. Specifically, existing methods can be divided into two popular supervisions, {\em i.e.}, strong~\cite{zeng2019graph,lin2021learning,qing2021temporal} and weak~\cite{wang2017untrimmednets,ju2023constraint,ju2020point,yudistira2022weakly}. Strong supervision gives precise boundary labels and category labels for training. There are two detailed pipelines: the top-down framework~\cite{shou2016temporal,shou2017cdc,gao2017turn,chao2018rethinking,lin2017single,xu2017r,tan2021relaxed,zhu2021enriching,wang2022rcl,xu2020g} pre-defines extensive anchors, adopts fixed-length sliding windows to produce initial proposals, then regresses to refine boundaries; the bottom-up framework~\cite{zhao2017temporal,lin2018bsn,lin2019bmn,vo2023aoe,zhao2020bottom,bai2020boundary} learns frame-wise boundary detectors for the boundary frames, then groups extreme frames or estimates action lengths for proposal generation. In addition, several works~\cite{gao2018ctap,liu2019multi,yang2020revisiting} used various fusion strategies to complement these frameworks. On the other hand, weak supervision trains without boundary labels to alleviate annotation costs. The video-level setting learns from category labels~\cite{paul2018w,ju2022distilling}, the CAS-based framework~\cite{liu2019completeness,ju2021adaptive,min2020adversarial,narayan2021d2,lee2019background,lee2021weakly,zhao2021soda} and attention-based framework~\cite{nguyen2018weakly,luo2021action,nguyen2019weakly,shi2020weakly,gao2022fine,he2022asm,huang2021foreground,luo2020weakly,ma2022weakly} have been well studied. To generate better results from CAS or attention, some studies~\cite{shou2018autoloc,liu2019weakly} improved post-processing. To balance cost and performance, some papers introduced single-frame annotations~\cite{ju2021divide,ma2020sf,lee2021learning,yang2021background,mettes2019pointly} or instance-number annotations~\cite{narayan20193c,xu2019segregated}. 

Nevertheless, all the above methods assume that action categories remain identical for training and testing, which is an over-simplification of real application scenarios, limiting practical uses of the vision system.

\vspace{0.1cm} 
{\noindent \bf Low-Shot Temporal Action Localization} considers more realistic scenarios: generalize TAL towards action categories that are unseen (zero-shot) or with several support samples (few-shot). Existing methods~\cite{ju2022prompting,nag2022zero,zhang2022ow,bao2022opental} most rely on foundational models pre-trained on large-scale image-caption pairs for help. Typically, E-Prompt~\cite{ju2022prompting} is the first to construct wide baselines with popular prompt tuning~\cite{Lester21,li21-prefixtuning} and vanilla temporal modeling. STALE~\cite{nag2022zero} explores the one-stage framework to further simplify usage. Although promising, all above methods meet two main challenges: (1) For category semantics, the definition may be vague, inaccurate, or incomplete. (2) For visual motions, temporal modeling may be insufficient. In this paper, for detailed category understanding, we design novel language descriptions from LLMs and vision-conditional prompt tuning; for clearer motion understanding, we introduce optical flows to provide explicit motion inputs.

\begin{figure*}[t]
\begin{center}
\includegraphics[width=\textwidth] {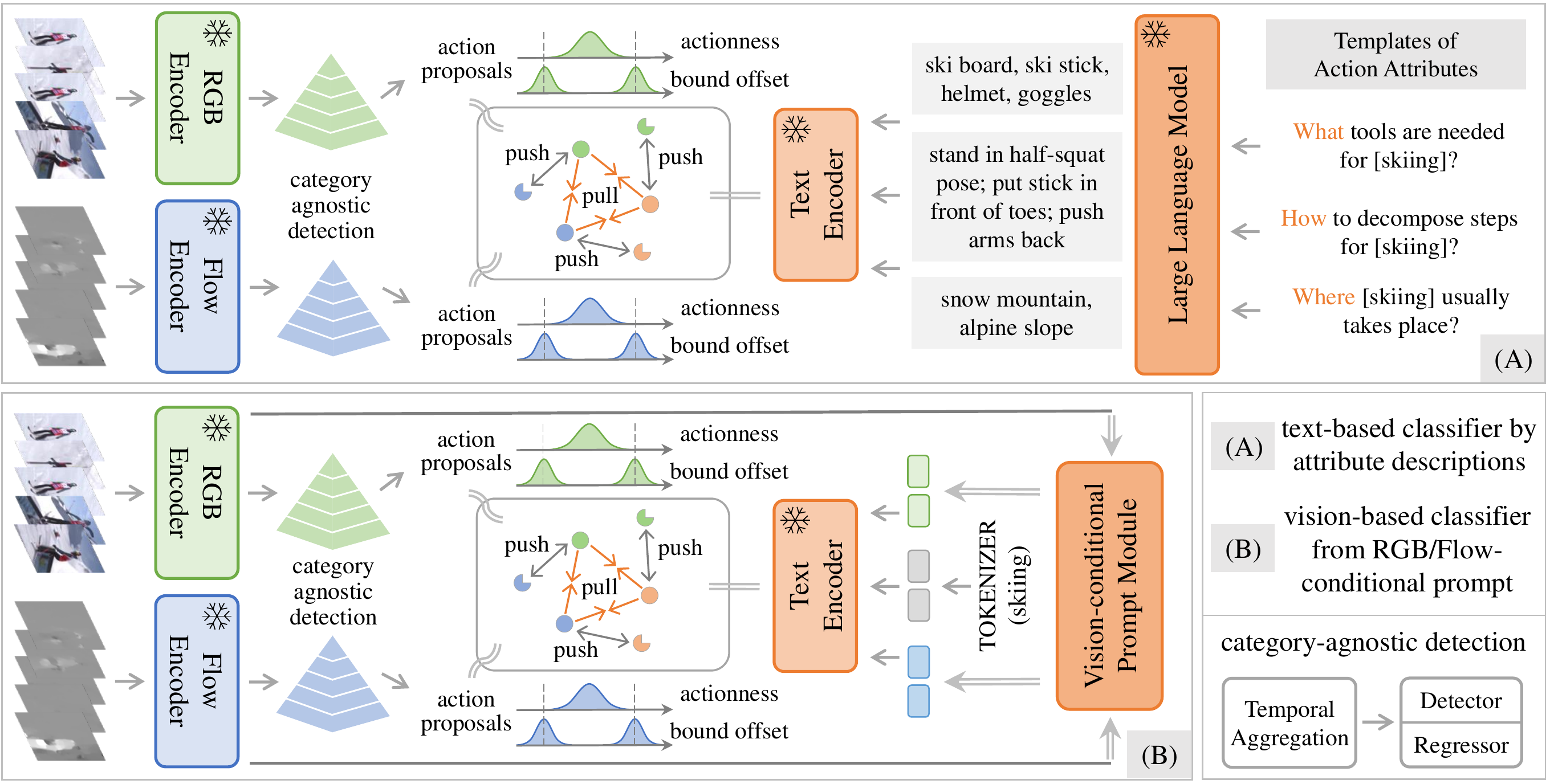}
\end{center}
\vspace{-0.25cm}
\caption{\textbf{Framework Overview.} Given one input video, we first encode RGB and Flow for appearance and motion embeddings, then localize category-agnostic action proposals (detect actionness and regress boundary). For open-vocabulary classification, we design (A) text-based classifiers using attribute descriptions from Large Language Models, and (B) vision-based classifiers using (RGB \& Flow)-conditional prompt tuning. Finally, the RGB-Flow-Text modalities are aligned for low-shot TAL.}
\vspace{0.15cm}
\label{fig:framework}
\end{figure*}

\section{Method}
\label{sec:method}
In this paper, our goal is to tackle open-vocabulary temporal action localization, by aligning multi-modal video representations, extracted from RGB, Flows, and texts. 
In Sec.~\ref{subsec:formulation}, we start by introducing problem scenarios. 
In Sec.~\ref{subsec:arch}, we describe the proposed architecture, consisting of two vital modules, namely, category-agnostic action proposal and open-vocabulary action classification via descriptions or visual instance prompts. 
Lastly, we detail the training and inference procedure in Sec.~\ref{subsec:training}.

\vspace{-0.2cm}
\subsection{Problem Scenario}  
\label{subsec:formulation}
\vspace{-0.1cm}
Assuming we are provided one untrimmed video dataset for training, {\em e.g.}, $\mathcal{D}_{\text{train}} = \{(\mathbf{V}_1, \mathbf{Y}_1), \dots, (\mathbf{V}_N, \mathbf{Y}_N)\}$, where $\mathbf{V}_i \in \mathbb{R}^{T_i \times H \times W \times 3}$ denotes one video sequence with $T_i$ frames, $\mathbf{Y}_i \in \mathbb{R}^{T_i \times \mathcal{C}_{\text{train}}}$ refers to the frame-wise annotation for action categories, that fall into one pre-defined vocabulary. 
Our goal is to train an open-vocabulary temporal action localization model, that can process testing videos from unseen categories, {\em i.e.}, to detect and classify actions beyond the seen ones at training time:
\begin{align}
    \mathbf{Y}_j = \Phi_{\text{open-TAL}}(\mathbf{V}_j) \in \mathbb{R}^{T_j \times \mathcal{C}_{\text{test}}}, 
    \quad 
    \mathbf{V}_j \sim \mathcal{D}_{\text{test}},
\end{align}
where $\mathbf{V}_j$ refers to a video sampled from the testing set. Under zero-shot settings, action categories for training (base) and evaluation (novel) are disjoint, {\em i.e.},~$\mathcal{C}_{\text{train}} \cap \mathcal{C}_{\text{test}} = \emptyset$. Under few-shot settings, training categories are included in the testing categories, {\em i.e.},~$\mathcal{C}_{\text{test}} \supset \mathcal{C}_{\text{train}}$.
As a comparison, in closed-set scenarios, categories for training and testing are identical, {\em i.e.},~$\mathcal{C}_{\text{test}} = \mathcal{C}_{\text{train}}$.

\vspace{-0.2cm}
\subsection{Architecture}
\label{subsec:arch}
\vspace{-0.1cm}
We propose a multi-modal low-shot action localization framework, containing two modules: category-agnostic proposal (in Sec.~\ref{sec:temp}), and open-vocabulary classification by descriptions or visual prompts~(in Sec.~\ref{sec:classifier}). \textbf{Note that}, both visual and textual encoders are pre-trained and kept frozen, thus resembling the downstream adaptations of foundation models~\cite{Radford21,Jia21,yao2021filip} effectively.

\subsubsection{Category-agnostic Action Proposal} 
\label{sec:temp}
As for the visual encoding, we here consider two widely-used modalities for video understanding, namely, RGB appearance and optical flow.

\vspace{0.2cm}
\noindent \textbf{RGB Frame Encoder.}
\hspace{1pt} Given one untrimmed video with $T$ frames, $\mathbf{V} = \{I_1, \dots, I_T\}$, we employ two types of RGB pre-trained encoders to obtain frame-wise features, {\em i.e.}, CLIP image~\cite{Radford21} and I3D RGB~\cite{carreira2017quo}. 
\begin{align}
    \mathbf{F}^{\mathrm{rgb}} = \Phi_{\text{CLIP-r}}(\mathbf{V}) \ \  \mathrm{or} \ \  \mathbf{F}^{\mathrm{rgb}} = \Phi_{\text{I3D-r}}(\mathbf{V}),
\end{align}
where $\Phi_{\text{CLIP-r}}(\cdot)$ is pre-trained on 400M image-text pairs, 
with rich description of visual appearances; 
$\Phi_{\text{I3D-r}}(\cdot)$ is pre-trained on Kinetics-400 for action recognition, with good representations of the temporal continuity. 
For simplicity, we denote the above features as $\mathbf{F}^{\mathrm{rgb}} \in \mathbb{R}^{T \times D_{\mathrm{rgb}}}$, $D_{\mathrm{rgb}}$ denotes the dimension of RGB features.

\vspace{0.2cm}
\noindent \textbf{Flow Encoder.} 
\hspace{1pt} We first use the TV-L1 algorithm~\cite{wedel2009improved} to compute optical flows from the RGB frame sequence, and then, feed them into the pre-trained I3D Flow encoder~\cite{carreira2017quo} to compute frame-wise motion features: 
\begin{align}
    \mathbf{F}^{\mathrm{flow}} =  \Phi_{\text{I3D-f}}(\mathbf{V}) \in \mathbb{R}^{T \times D_{\mathrm{flow}}},
\end{align}
where $\Phi_{\text{I3D-f}}(\cdot)$ is pre-trained on Kinetics-400, thus containing rich motion details, $D_{\mathrm{flow}}$ denotes the dimension of the Flow features.

\vspace{0.2cm}
\noindent \textbf{Temporal Aggregation.}
\hspace{1pt} Given visual features ($\mathbf{F}^{\mathrm{rgb}}$ \& $\mathbf{F}^{\mathrm{flow}}$) from the frozen encoders, we pass them through one temporal aggregation module $\Phi_{\text{temp}}(\cdot)$, consisting of standard Transformer encoder layers, for better action modeling. As the action duration varies frequently, following recent methods~\cite{dai2022ms,zhang2022actionformer}, we construct a multi-scale pyramid structure for $\Phi_{\text{temp}}$, formulated as: 
\begin{align}
    \mathbf{F}^{*} = \{\mathbf{F}_1^{*}, ...,  \mathbf{F}_L^{*} \} =  \Phi_{\text{temp}}(\mathbf{F}^{*})
    \in \mathbb{R}^{T' \times D_{*}},
\end{align}
where $*$ refers to the RGB or Flow modality. The Transformer pyramid covers $L$ layers, and each layer consists of Multi-head Self-attention, Layer Norm, and MLPs. We perform ``$2\times$'' down-sampling between adjacent layers, to ensure that deeper layers focus on long actions, while shallow layers handle short actions. As a result, we obtain $\mathbf{F}_i^{*} \in \mathbb{R}^{T/{2^{i}} \times D_{*}}$ for the $i$-th layer features.

\vspace{0.2cm}
\noindent \textbf{Category-agnostic Proposals.} 
\hspace{1pt} Here, we design one temporal action localizer, containing one detector $\Phi_{\mathrm{det}}(\cdot)$ and one regressor $\Phi_{\mathrm{reg}}(\cdot)$ in parallel, to produce category-agnostic action proposals. Concretely, when taking multi-scale pyramid features $\mathbf{F}^{*}$ as input, $\Phi_{\mathrm{det}}(\cdot)$ predicts the frame-level action probability, {\em i.e.}, performing the binary classification for action and background; while $\Phi_{\mathrm{reg}}(\cdot)$ regresses the left offset and right offset to the nearest action instance for each frame, thus refining the boundary. 
\begin{equation}   \label{eq:detector}
{
\widehat{\mathbf{p}}_{\mathrm{det}}^{*} = \Phi_{\mathrm{det}}(\mathbf{F}^{*}) \in \mathbb{R}^{T'},
\ \
\widehat{\mathbf{P}}_{\mathrm{reg}}^{*} = \Phi_{\mathrm{reg}}(\mathbf{F}^{*}) \in \mathbb{R}^{T' \times 2}.
}
\end{equation}

Structurally, both the detector and the regressor are composed of 1D convolutional networks. Hereafter, by further post-processing $\widehat{\mathbf{p}}_{\mathrm{det}}$ and $\widehat{\mathbf{P}}_{\mathrm{reg}}$, we could obtain $B$ action proposals $\{{b_i}\}_{i=1}^B$. And then, we compute visual embeddings for these category-agnostic proposals by taking the mean pooling of RGB or Flow features within the proposal intervals.
\begin{align}
    \mathbf{f}^{*}_i = \Phi_{\text{pool}}(\mathbf{F}^{*}[b_i]) \in \mathbb{R}^{D_{*}}.
\end{align}

\textbf{Note that}, we do not claim novelty or contribution on category-agnostic action detection~\cite{Lin18,zhao2020bottom,tan2021relaxed,zhang2022actionformer,lin2021learning}, which has been widely-studied in the community. And our method is flexible to these off-the-shelf detectors.

\subsubsection{Open-vocabulary Action Classification}
\label{sec:classifier}
In this section, we introduce the procedure for classifying action proposals towards both base and novel categories~(only requested by user at inference time). To produce such open-vocabulary classifiers, the key is to use a pre-trained text encoder of vision-language foundation models. Specifically, one vanilla solution~\cite{Radford21} is to combine category names with handcrafted prompts, {\em e.g.}, ``one video of \{category\}'', then feed into the CLIP text encoder to generate action classifier. 
However, such a paradigm suffers from lexical ambiguities, for example, “fencing” can either refer to “the game of stabbing with slender steel swords in protective clothing” or “enclose or separate something with paling or hedgerow”, directly encoding vanilla category names is unable to distinguish these concepts. Here, we present two novel strategies to enhance the discriminative power of generated classifiers with: detailed language descriptions or vision-conditioned prompting.

\vspace{0.2cm}
\noindent \textbf{Classifier construction by language descriptions. } 
Here, we consider to decompose actions into ``attribute'' that specifies one category from various aspects, hence enriching the discriminative power of constructed classifiers. As shown in Figure~\ref{fig:framework} (A), for $C$ action category names (denoting as $\mathcal{C}_{\mathrm{name}}$), we prompt one large-scale language model~(LLMs), for example, GPT-3~\cite{Brown20} or OPT~\cite{zhang2022opt}, with three attribute templates, to obtain detailed descriptions of salient objects, event fields, and motion interactions for query actions:
\begin{align}
\mathbf{M}_{\mathrm{desc}} &= \Phi_{\text{LLM}}(\Phi_\mathrm{attr}[\mathcal{C}_{\mathrm{name}}]),
\end{align}
where three attribute templates $\Phi_\mathrm{attr}$ are ``what tools are needed for [action]?'', ``where [action] usually takes place?'', and ``how to decompose steps for [action]?'', respectively. Note that, here we only use some straight-forward prompt templates, while other templates may also bring similar attribute descriptions.

With the attribute descriptions $\mathbf{M}_{\mathrm{desc}}$, the classifier embeddings for $C$ action categories can thus be generated using the CLIP text encoder:
\begin{align}  \label{description}
\mathbf{F}^{\mathrm{text}} = \Phi_{\text{CLIP-t}}( \Phi_\textsc{tokenise}(\mathbf{M}_{\mathrm{desc}})) \in \mathbb{R}^{C \times D_{\mathrm{text}}},
\end{align}
where $\Phi_\textsc{tokenise}(\cdot)$ refers to the language tokenizer that converts words into vectors, and $D_{\mathrm{text}}$ refers to the dimension of the textual features.

\vspace{0.2cm}
\noindent \textbf{Classifier construction by conditioning on video instance.} 
\hspace{1pt} For the cases where comprehensive text descriptions are challenging to acquire, for instance, a good description for gymnastics with intricate pose patterns is prohibitively long, we consider one alternative way for lexical disambiguation. In particular, we design a vision-conditional prompt module $\Phi_{\mathrm{prmp}}(\cdot)$, as shown in Figure~\ref{fig:framework} (B). For any given video, $\Phi_{\mathrm{prmp}}(\cdot)$ takes its RGB and Flow features, {\em i.e.}, $\mathbf{F}^{\mathrm{rgb}}$ and $\mathbf{F}^{\mathrm{flow}}$, as inputs, and outputs $K$ concrete prompt vectors $\mathbf{M}_{\mathrm{cond}}$. 
\begin{align}
    \mathbf{M}^{*}_{\mathrm{cond}} = \Phi_{\text{prmp}}(\mathbf{F}^{*}) \in \mathbb{R}^{K \times D_{\mathrm{text}}}.
\end{align}
where $\Phi_{\mathrm{prmp}}(\cdot)$ refers to a trainable module~(can be either MLPs or Transformer encoder), and $\mathbf{M}^*_{\mathrm{cond}}$ denotes the instance-specific prompt vectors, containing rich visual details from the RGB or Flow stream.

Hereafter, we can generate the classifier embeddings for ${C}$ actions, {\em i.e.}, prepending\,/\,appending $\mathbf{M}^{*}_{\mathrm{cond}}$ with category name tokens, then inputting all these tokens into the CLIP text encoder. 
\begin{align}  \label{conditional}
\mathbf{F}^{\mathrm{text}} = \Phi_{\text{CLIP-t}}(\mathbf{M}^{*}_{\mathrm{cond}},\, \Phi_\textsc{toke}(\mathcal{C}_{\mathrm{name}}),\, \mathbf{M}^{*}_{\mathrm{cond}}),
\end{align}
where $\mathbf{F}^{\mathrm{text}} \in \mathbb{R}^{C \times D_{\mathrm{text}}}$ refers to the classifier embeddings. At training time, the gradients can flow through the frozen encoder $\Phi_{\text{CLIP-t}}(\cdot)$ to only optimize the conditional prompt module $\Phi_{\mathrm{prmp}}(\cdot)$.

In zero-shot scenarios, $\Phi_{\mathrm{prmp}}(\cdot)$ is trained on seen (base) categories, then directly applied to test videos of novel categories, outputting vision-conditional, task-specific prompt vectors $\mathbf{M}_{\mathrm{cond}}$; while in the few-shot scenarios, $\mathbf{M}_{\mathrm{cond}}$ can be category-specific to learn from the support video exemplars of novel categories. Intuitively, $\Phi_{\mathrm{prmp}}(\cdot)$ can be seen as implicitly decomposing actions into ``visual attributes'', enriching the classifier obtained from only encoding category names.

\vspace{0.2cm}
\noindent \textbf{Cross-modal alignment.}
For one video, the textual stream ends up with the category-wise features $\mathbf{F}^{\mathrm{text}} = \{\mathbf{f}_1^{\mathrm{text}}, \dots, \mathbf{f}_{C}^{\mathrm{text}} \} \in \mathbb{R}^{C \times D_{\mathrm{text}}}$; while the visual stream ends up with RGB features $\mathbf{f}^{\mathrm{rgb}}\in \mathbb{R}^{D_{\mathrm{rgb}}}$ and Flow features $\mathbf{f}^{\mathrm{flow}} \in \mathbb{R}^{D_{\mathrm{flow}}}$ for action proposals.

To alignment between language with appearance or motion, we adopt a lightweight module~($\Phi_{\mathrm{align}}(\cdot)$) to map RGB-Flow-Text embeddings into one shared space.
\begin{equation}   \label{eq:aligner}
{
\mathbf{h}^{\circ} = \Phi_{\mathrm{align}}(\mathbf{f}^{\circ}) \in \mathbb{R}^{D_{\mathrm{align}}},
}
\end{equation}
where $\circ$ refers to any of the three modalities, and $D_{\mathrm{align}}$ refers to the aligned dimension of the embedding space. Architecture-wise, $\Phi_{\mathrm{align}}(\cdot)$ is flexible to off-the-shelf networks, such as MLPs or Transformer encoder.

\vspace{-0.2cm}
\subsection{Training and Inference}
\label{subsec:training}
\vspace{-0.1cm}
Given one batch of (RGB, Flow, Text) training pairs, the visual stream ends up with RGB and Flow embeddings for action proposals, namely, $\mathbf{h}^{\mathrm{rgb}}$ and $\mathbf{h}^{\mathrm{flow}}$; while the textual stream ends up with classifier embeddings $\{\mathbf{h}_1^{\mathrm{text}}, \dots, \mathbf{h}_{C}^{\mathrm{text}} \}$. In the following, we describe the optimization details for detection and classification.

\vspace{0.2cm}
\noindent \textbf{Category-agnostic Proposal.} 
\hspace{1pt} Following previous methods~\cite{Lin18,tan2021relaxed}, to supervise the generation of action proposals, we adopt the weighted cross-entropy loss for action proposal~\cite{lin2019bmn,lin2018bsn} and use the DIoU loss $\mathcal{L}_{\mathrm{reg}}$ for distance regression~\cite{zheng2020distance}. Formally, 
\begin{equation}
{
\mathcal{L}_{\mathrm{det}} = \sum _{t\in \Omega^{+}} \mathcal{H}({p}_t^{*}, {\widehat{p}}_t^{*}) +  \gamma \sum _{t\in \Omega^{-}} \mathcal{H}({p}_t^{*}, {\widehat{p}}_t^{*}),
}
\end{equation}
where ${{p}_t} \in \{0,1\}$ and ${{\widehat{p}}_t} \in [0,1]$ are the label and probability for category-agnostic actions at $t$-th timestamp; $\mathcal{H}$ and $\gamma$ are the cross-entropy and the balancing weight; $\Omega^{+}$ and $\Omega^{-}$ are positive sets and negative sets. 
\begin{equation}
{\mathcal{L}_{\mathrm{reg}} = 1 - \mathrm{IoU} + \mathcal{D}(\widehat{\mathbf{P}}_{\mathrm{reg}}^{*}, \mathbf{P}_{\mathrm{reg}}^{*}),
}
\end{equation}
where $\mathrm{IoU}$ is the intersection over union between predicted proposals and ground-truth actions. $\mathbf{P}_{\mathrm{reg}}$ refers to the regression ground-truth, and $\mathcal{D}$ refers to the normalized euclidean distance.

\vspace{0.2cm}
\noindent \textbf{Open-vocabulary Classification.}
\hspace{1pt} After calculating action proposals, we aim to encourage the paired (RGB, Flow, Text) embeddings to emit the highest similarity score among others, we use the infoNCE loss to supervise tri-modal alignment in the shared space, which can be written as follows:
\begin{align}
\mathcal{L}_{\mathrm{cls}} = &- \sum_i \big( \log \frac{\exp( \mathcal{S}(\mathbf{h}_{i}^{\mathrm{*}}, \mathbf{h}_{i}^{\mathrm{text}}) / \tau)}{\sum\limits_{j} \exp(\mathcal{S}(\mathbf{h}_{i}^{\mathrm{*}}, \mathbf{h}_{j}^{\mathrm{text}}) / \tau)} \big),
\end{align}
where $\mathcal{S}$ refers to the cosine similarity, and $\tau$ is one temperature parameter. Using the well-aligned multi-modal features, we can naturally achieve open-vocabulary classification, through evaluating the cosine similarity between the textual and visual modalities.

\vspace{0.2cm}
\noindent \textbf{Total Loss.}
\hspace{1pt} During training, we freeze encoders for RGB-Flow-Text, 
and jointly optimize the detector, regressor, vision-conditional prompt module, and corss-modal aligner. Using two balancing ratios ($\lambda_1$ and $\lambda_2$), the total optimization loss $\mathcal{L}_{\mathrm{all}}$ can be formulated: 
\begin{align}
  \mathcal{L}_{\mathrm{all}} = \mathcal{L}_{\mathrm{det}} + \lambda_1\mathcal{L}_{\mathrm{reg}} + \lambda_2\mathcal{L}_{\mathrm{cls}}.
\end{align}

\vspace{0.1cm}
\noindent \textbf{Inference.}
\hspace{1pt} At testing time, for one given video, we could compute the frame-level action probability $\widehat{\mathbf{p}}_{\mathrm{det}}^{*}$ from the binary classifier, and the frame-level boundary offset $\widehat{\mathbf{P}}_{\mathrm{reg}}^{*}$ from the boundary regressor. For category-agnostic proposal, we threshold $\widehat{\mathbf{p}}_{\mathrm{det}}^{*}$ through $\theta_{loc}$, concatenate consecutive snippets as action proposals, utilize $\widehat{\mathbf{P}}_{\mathrm{reg}}^{*}$ for boundary refinement, and eliminate proposal redundancy with soft non-maximum suppression (NMS). For low-shot classification, we calculate cosine similarity between category textual embeddings and proposal visual embeddings, then only retain the category with probabilities greater than threshold $\theta_{cls}$. 

\vspace{-0.3cm}
\section{Experiments}
We experiment on two public datasets across four data splits. In Sec.~\ref{subsec:ablation} and~\ref{subsec:choices}, we validate the effectiveness of each component. In Sec.~\ref{subsec:SOTA}, we compare with state-of-the-art low-shot methods. In Sec.~\ref{subsec:qualitative}, we visualize the localization results.

\vspace{-0.2cm}
\subsection{Datasets \& Metrics}   
\label{subsec:implementation}
\noindent {\bf THUMOS14~\cite{jiang2014thumos}} has $413$ untrimmed videos from $20$ categories, with an average of $15$ instances per video, and $20$ videos per category. 
\noindent {\bf ActivityNet1.3}~\cite{caba2015activitynet} covers 20k videos from $200$ categories, with an average of $1.5$ instances per video, and $100$ videos per category.

\vspace{0.2cm}
\noindent \textbf{Splits.}
Following literature~\cite{ju2022prompting,nag2022zero}, we adopt two types of splits for zero-shot scenarios. 
{\bf The 75:25 split}: train on 75\% base categories and test on 25\% novel categories.
{\bf The 50:50 split}: train on 50\% base categories and test on 50\% novel categories.
The final results are calculated by averaging $10$ random splits. 

While for few-shot scenarios, as we are not aware of any existing benchmarks, we initiate the $N$-shot evaluation, {\em i.e.}, sample $N$ videos from the training set for each novel category, to form the few-shot support set, and then measure on the standard testing set.

\vspace{0.2cm}
\noindent \textbf{Metrics.} To evaluate localization performance, we report mean Average Precision (mAP) under different intersections over union (IoU) thresholds, following standard protocols. To evaluate classification performance, we report the TOP1 accuracy. Note that one proposal is regarded as positive only if both the category prediction is correct and the IoU exceeds set thresholds.

\vspace{-0.3cm}
\subsection{Implementation Details}   
Our framework is implemented with PyTorch, and all experiments are conducted on one 24G GeForce RTX 3090 GPU. On all datasets, the models are optimized with Adam, using a learning rate of $10^{-4}$, and a batch size of $32$ videos. We warm up the model in the first $5$ epochs for better convergence, and continue to train $45$ epochs for full optimization. To deal with the large variety in video durations, we pad all videos with zeros to $T$ frames. $T$ is set to $192$ on ActivityNet1.3, and $2304$ on THUMOS14. For temporal resolution, we take $16$ consecutive frames as one basic input unit, and the stride of sliding windows is set to $4$ frames, following the literature~\cite{zhang2022actionformer,liu2019completeness,ju2022prompting}. For spatial resolution, we use center crop on each video frame to get $224\times224$ image. For the text stream, we employ the CLIP text encoder $D_{\mathrm{text}}=512$. For the Flow encoder, we utilize the I3D network $D_{\mathrm{flow}}=1024$. For the RGB stream, we explore two solutions: the CLIP image encoder $D_{\mathrm{rgb}}=512$ and the I3D network $D_{\mathrm{rgb}}=1024$. Architecture-wise, both the CLIP image and text encoders are ViT-B/16. We use GPT-3~\cite{Brown20} for $\Phi_{\text{LLM}}(\cdot)$; adopt fully convolutional networks for detector $\Phi_{\mathrm{det}}$ and regressor $\Phi_{\mathrm{reg}}(\cdot)$; employ MLPs for vision-conditional prompt mudule $\Phi_{\mathrm{prmp}}(\cdot)$.

All hyperparameters are set by the grid search: balancing ratios $\lambda_1=\lambda_2=1$, the temperature $\tau=0.07$, detection threshold $\theta_{cls}=0.85$, classification threshold $\theta_{loc}= 0.05$, and the soft-NMS threshold is set to $0.5$. The dimension of multi-modal shared space is $D_{\mathrm{align}}=1024$, and the pyramid layer $L=6$.

\begin{table}[t]
\footnotesize
\setlength\tabcolsep{5pt}
\centering
\begin{tabular}{c|c|cc|c|c}
\toprule
\multirow{2}{*}{Text} & \multirow{2}{*}{Modality} & \multicolumn{2}{c|}{mAP@IoU} & \multirow{2}{*}{\begin{tabular}[c]{@{}c@{}}AVG\\ (0.3-0.7)\end{tabular}} & \multirow{2}{*}{ACC} \\ \cline{3-4}
 &  & 0.3 & 0.7 &  &  \\ \midrule
\multirow{2}{*}{$\mathcal{C}_{\mathrm{name}}$} & RGB & 33.8 & 6.2 & 20.7 & 79.3 \\
 & RGB+Flow & 46.8 & 12.9 & 31.5 & 79.4 \\ \hline
\multirow{2}{*}{$\mathbf{M}_{\mathrm{desc}}$} & RGB & 47.4 & 11.7 & 31.2 & 85.3 \\
 & RGB+Flow & 54.3 & 16.7 & 37.8 & 86.7 \\ 
\bottomrule
\end{tabular}
\caption{\textbf{Text-based classifiers from language descriptions.} For classifier generation, decomposing vanilla category names into attribute descriptions through LLMs, can provide additional action details to enrich discriminative information for classifiers, thus alleviating lexical confusion.}
\vspace{0.1cm}
\label{tab:caption}
\end{table}

\begin{table}[t]
\footnotesize
\setlength\tabcolsep{5pt}
\centering
\begin{tabular}{c|c|c|c|c}
\toprule
\multirow{2}{*}{Descriptions} & \multirow{2}{*}{Fusion} &  \multirow{2}{*}{Modality} &
\multirow{2}{*}{\begin{tabular}[c]{@{}c@{}}AVG\\ (0.3-0.7)\end{tabular}}  &
\multirow{2}{*}{ACC} \\ 
 &  &  &  \\ \hline
How & \multirow{3}{*}{--} & \multirow{6}{*}{\begin{tabular}[c]{@{}c@{}}RGB\\ + \\Flow\end{tabular}} & 33.3 & 80.3 \\
Where & & & 35.0 & 84.0 \\
What & & & 35.5 & 84.2 \\ \cline{0-1} \cline{4-5}
\multirow{3}{*}{\begin{tabular}[c]{@{}c@{}}How+What\\ +Where\end{tabular}}
 & average & & 34.6 & 83.1 \\ 
 & concat & & 36.8 & 84.3 \\
 & weight & & 37.8 & 86.7 \\ 
 \bottomrule
\end{tabular}
\caption{\textbf{Descriptions of attribute questions.} We adopt ``what, where, how'' to prompt LLMs for salient objects, event fields, and motion interactions, respectively. The weighted fusion of these three descriptions brings the best classifier.}
\vspace{-0.2cm}
\label{tab:description}
\end{table}

\begin{table}[t]
\footnotesize
\setlength\tabcolsep{5pt}
\centering
\begin{tabular}{c|c|cc|c|c}
\toprule
\multirow{2}{*}{Prompt} & \multirow{2}{*}{Modality} & \multicolumn{2}{c|}{mAP@IoU} & \multirow{2}{*}{\begin{tabular}[c]{@{}c@{}}AVG\\ (0.3-0.7)\end{tabular}} & \multirow{2}{*}{ACC} \\ \cline{3-4}
 &  & 0.3 & 0.7 &  &  \\ \midrule
\multirow{2}{*}{\xmark} & RGB & 33.8& 6.2 & 20.7 & 79.3 \\
 & RGB+Flow & 46.8 & 12.9 & 31.5 & 79.4 \\ \hline
\multirow{2}{*}{$\mathbf{M}_{\mathrm{rand}}$} & RGB & 39.7 & 9.7 & 26.0 & 81.7 \\
 & RGB+Flow & 50.3 & 16.1 & 35.5 & 82.3 \\ \hline 
\multirow{2}{*}{$\mathbf{M}_{\mathrm{cond}}$} & RGB & 46.8 & 11.5 & 30.7 & 84.0 \\
 & RGB+Flow & 56.3 & 17.3 & 39.2 & 87.6 \\ 
\bottomrule
\end{tabular}
\caption{\textbf{Vision-conditional prompt module.} Compared to the prompt tuning in the text stream ($\mathbf{M}_{\mathrm{rand}}$), {\em i.e.}, prompt vectors are randomly initialized and learnt on seen categories, our vision-conditional prompts ($\mathbf{M}_{\mathrm{cond}}$) characterize more visual details, thus strengthening generalization.}
\vspace{0.1cm}
\label{tab:prompt}
\end{table}

\begin{table}[t]
\footnotesize
\setlength\tabcolsep{6pt}
\centering
\begin{tabular}{c|c|c|c}
\toprule 
\multirow{2}{*}{Modality} & \multirow{2}{*}{\begin{tabular}[c]{@{}c@{}}Oracle AVG\\ (0.3-0.7)\end{tabular}} & \multirow{2}{*}{\begin{tabular}[c]{@{}c@{}}AVG\\ (0.3-0.7)\end{tabular}} & \multirow{2}{*}{ACC}\\ 
 &  &  &  \\ \midrule
RGB-Text\,$\dag$ & 30.5 & 29.2 & 97.9  \\
RGB-Text\,$\ddag$ & 40.4 & 30.1 & 83.8  \\
Flow-Text & 46.5 & 32.3 & 76.7\\ \hline
RGB-Text-Flow\,$\dag$ & 47.4 & 43.7 & 98.0 \\ 
RGB-Text-Flow\,$\ddag$ & 48.9 & 39.2 & 87.6 \\ 
\bottomrule
\end{tabular}
\caption{\textbf{RGB-Flow-Text tri-modal alignment.} `Oracle' refers to detection using ground-truth category labels. $\dag$ and $\ddag$ refer to using RGB encoders from CLIP or I3D. By effectively aligning to Text, RGB and Flow encoders can handle low-shot tasks. The tri-modal alignment shows the best performance.}
\vspace{-0.2cm}
\label{tab:ablatalign}
\end{table}

\vspace{-0.2cm}
\subsection{Ablation Study}    
\label{subsec:ablation}
We here ablate key components to evaluate their effectiveness. Unless otherwise stated, experiments are conducted under the 75:25 zero-shot split on THUMOS14, using the I3D encoders for both RGB and Flow modalities, CLIP encoder for text modality.

\vspace{0.2cm}
\noindent \textbf{Text-based classifiers from detailed language descriptions.} 
To alleviate lexical confusion for vanilla category names $\mathcal{C}_{\mathrm{name}}$, we decompose actions into attribute descriptions $\mathbf{M}_{\mathrm{desc}}$, with the help of Large-scale Language Models (GPT-3). Table~\ref{tab:caption} compares the performance resulting from these two options.

Comparing to only using category names for action classification, incorporating detailed descriptions could enrich discriminative information for classifiers. As a result, $\mathbf{M}_{\mathrm{desc}}$ could bring around 10.5\% average mAP gains and 6.0\% accuracy gains over $\mathcal{C}_{\mathrm{name}}$, proving the effectiveness of category completion. On the one hand, attribute descriptions specify one category name from various aspects, hence giving additional action details. On the other hand, the descriptions from LLMs avoid manually spending time to search external knowledge sources, thus they are also efficient for use.

\vspace{0.2cm}
\noindent \textbf{Optimal language descriptions.} 
\hspace{1pt} For descriptions $\mathbf{M}_{\mathrm{desc}}$ from LLMs, there are various prompt templates available. For full attribute descriptions of various aspects, we use three types of templates, {\em i.e.}, ``what tools?'', ``where takes place?'', and ``how to decompose steps?'', to obtain salient objects, event fields, and motion interactions. Table~\ref{tab:description} evaluates their effectiveness.

Single prompt only brings trivial results, and ``what tools?'' performs best among the three attributes. To integrate the diversity, we propose three simple strategies, {\em i.e.}, concatenate descriptions then text encoding, average their text encodings, and weighted sum the text encodings. Overall, prompt fusion brings better performance, {\em e.g.}, 2.8\% average mAP gains and 2.6\% accuracy gains over any single prompt. We finally choose weighted sum for its best results.

\begin{table}[t]
\footnotesize
\setlength\tabcolsep{4.5pt}
\centering
\begin{tabular}{c|c|cc|c|c}
\hline
\multirow{2}{*}{Position} & \multirow{2}{*}{Modality} & \multicolumn{2}{c|}{mAP@IoU} & \multirow{2}{*}{\begin{tabular}[c]{@{}c@{}}AVG\\ (0.3-0.7)\end{tabular}} & \multirow{2}{*}{ACC} \\ \cline{3-4}
 &  & 0.3 & 0.7 &  &  \\ \hline
Input & \multirow{2}{*}{RGB} & 46.8 & 11.5 & 30.7 & 84.0 \\
Output & & 46.3 & 11.4 & 30.4 & 84.4 \\ \hline
Input & \multirow{2}{*}{RGB+Flow} & 56.3 & 17.3 & 39.2 & 87.6  \\
Output &  & 56.1 & 17.6 & 39.2 & 87.9 \\ \hline
\end{tabular}
\caption{\textbf{Usage positions of vision-conditional prompt vectors $\mathbf{M}_{\mathrm{cond}}$.} For the CLIP text encoder, putting $\mathbf{M}_{\mathrm{cond}}$ into the input or the output produces similar results.}
\vspace{0.1cm}
\label{tab:position}
\end{table}

\begin{figure}[t]
\begin{center}
\includegraphics[width=0.38\textwidth] {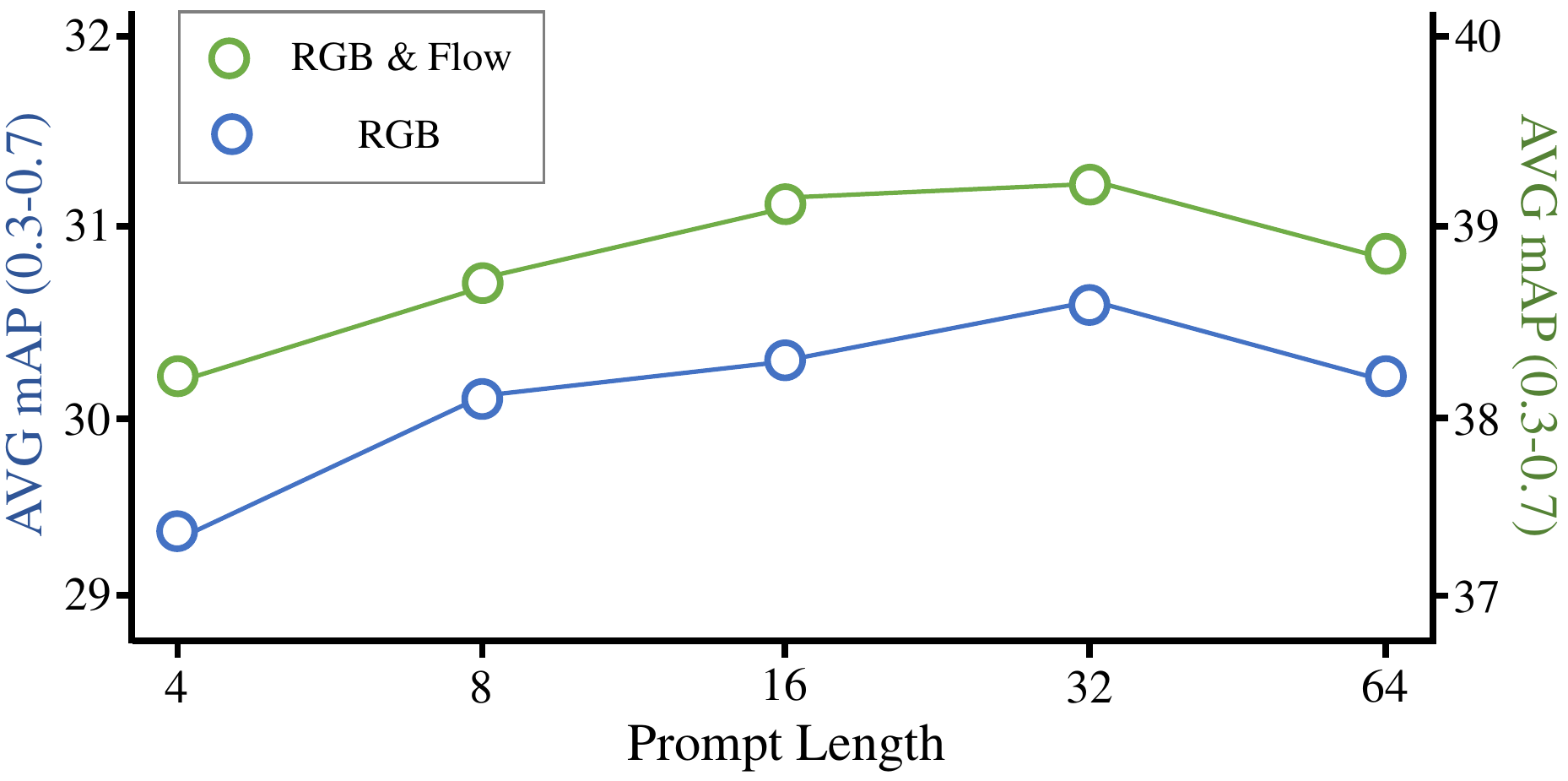}
\end{center}
\vspace{-0.4cm}
\caption{\textbf{Effect of prompt length.} The variance in prompt length has only slight effects on localization performance.}
\label{fig:length}
\end{figure}

\vspace{0.2cm}
\noindent \textbf{Vision-conditional prompt tuning.} 
\hspace{1pt} For the case on building classifiers with video-conditioned prompts, {\em i.e.}, feed RGB and Flow embeddings into the prompt module $\Phi_{\text{prmp}}(\cdot)$, to generate vision-conditional prompt vectors $\mathbf{M}_{\mathrm{cond}}$. Table~\ref{tab:prompt} compares $\mathbf{M}_{\mathrm{cond}}$ with $\mathbf{M}_{\mathrm{rand}}$, where the latter refers to the popular prompt tuning~\cite{zhu2022prompt,ju2022prompting,Lester21}, with the prompt vectors being randomly initialized and learnt on the base (seen) categories.

Comparing to the case without prompt, $\mathbf{M}_{\mathrm{rand}}$ has shown considerable improvements, for both detection and classification. While our vision-conditional prompt vectors $\mathbf{M}_{\mathrm{cond}}$ further boost the performance. For instance, when using only RGB modality, 2.3\% accuracy gains and 4.7\% average mAP gains over $\mathbf{M}_{\mathrm{rand}}$. Our $\Phi_{\text{prmp}}(\cdot)$ enables the prompt vectors $\mathbf{M}_{\mathrm{cond}}$ to be instance-specific, that enables to characterize rich visual details, effectively complementing the information acquired from only encoding the action category names.

Moreover, the prompt vectors learned from RGB-Flow dual modalities significantly outperform those from uni-modality. This is because, in terms of describing visual action details, the RGB modality focuses more on appearance or context; while the Flow modality mainly consists of motion information, the dual-modal prompt vectors could complement each other.

\begin{table}[t]
\footnotesize
\setlength\tabcolsep{7pt}
\centering
\begin{tabular}{c|c|c|c}
\toprule 
\multirow{2}{*}{Modality} & \multirow{2}{*}{Share} & \multirow{2}{*}{\begin{tabular}[c]{@{}c@{}}AVG\\ (0.3-0.7)\end{tabular}} & \multirow{2}{*}{ACC} \\ 
 &  &  \\ \midrule
RGB-Text-Flow\,$\dag$ & \multirow{2}{*}{No} & 43.7 & 98.0 \\ 
RGB-Text-Flow\,$\ddag$ & & 39.2 & 87.6 \\ \hline
RGB-Text-Flow\,$\dag$ & \multirow{2}{*}{Yes} & 33.5 & 83.3 \\
RGB-Text-Flow\,$\ddag$ & & 36.3 & 83.6 \\
\bottomrule
\end{tabular}
\caption{\textbf{Shared visual backbones for RGB and Flow.} $\dag$ and $\ddag$ refer to utilizing the RGB encoders from CLIP or I3D. `Oracle' means detection using ground-truth category labels. The shared backbones can harm performance somewhat.}
\vspace{0.3cm}
\label{tab:share}
\end{table}

\begin{table}[t]
\footnotesize
\setlength\tabcolsep{3.5pt}
\centering
\begin{tabular}{c|c|cc|c|c}
\toprule 
\multirow{2}{*}{Training} & \multirow{2}{*}{Modality} &\multicolumn{2}{c|}{mAP@IoU} & \multirow{2}{*}{\begin{tabular}[c]{@{}c@{}}AVG\\ (0.3-0.7)\end{tabular}} & \multirow{2}{*}{ACC} \\ \cline{3-4}
 & & 0.3 & 0.7 &  &  \\ \midrule
\multirow{1}{*}{Freeze} & \multirow{2}{*}{RGB+Flow} & 56.3 & 17.3 & 39.2 & 87.6 \\ 
\multirow{1}{*}{End-to-End} & & 41.7 & 12.8 & 29.3 & 71.3 \\ 
\bottomrule
\end{tabular}
\caption{\textbf{Optimization modes.} Comparing to freezing all encoders, end-to-end fine-tuning the entire model brings numerous parameters that tend to specialize on the base categories, thereby damaging the model generalization.}
\vspace{-0.1cm}
\label{tab:fixend}
\end{table}

\vspace{0.2cm}
\noindent \textbf{Effectiveness of optical flow.} 
Table~\ref{tab:ablatalign} validates the efficacy of multi-modal alignment. For evaluation metrics, we also report the Oracle mAP using GT category labels, to decouple proposal and classification.

\begin{table*}[t]
\footnotesize
\setlength\tabcolsep{5.5pt}
\centering
\vspace{0.08cm}
\begin{tabular}{c|c|c|ccccccccccc}   
\toprule 
\multirow{2}{*}{} & \multirow{2}{*}{} & \multirow{2}{*}{} & \multicolumn{6}{c}{THUMOS14~(mAP@IoU)} &  & \multicolumn{4}{c}{ActivityNet1.3~(mAP@IoU)} \\ \cmidrule{4-9} \cmidrule{11-14}
Setting & Method & Modality & 0.3 & 0.4 & 0.5 & 0.6 & 0.7 & AVG &  & 0.5 & 0.75 & 0.95 & AVG \\ \midrule
\multirow{7}{*}{\begin{tabular}[c]{@{}c@{}}Closed-Set\\ \\ 100\% Seen\\ 0\% Unseen\end{tabular}}
& \multicolumn{1}{c|}{TALNET~\cite{Chao18}} & \multicolumn{1}{c|}{RGB+Flow} & 53.2 & 48.5 & 42.8 & 33.8 & 20.8 & 39.8 &  & 38.2 & 18.3 & 1.3 & 20.2 \\
& \multicolumn{1}{c|}{BSN~\cite{Lin18}} & \multicolumn{1}{c|}{RGB+Flow} & 53.5 & 45.0 & 36.9 & 28.4 & 20.0 & 36.8 & & 46.5 & 30.0 & 8.0 & 30.0 \\
& \multicolumn{1}{c|}{BUTAL~\cite{zhao2020bottom}} & \multicolumn{1}{c|}{RGB+Flow} & 53.9 & 50.7 & 45.4 & 38.0 & 28.5 & 43.3 & & 43.5 & 33.9 & {\bf 9.2} & 30.1 \\
& \multicolumn{1}{c|}{A2NET~\cite{yang2020revisiting}} & \multicolumn{1}{c|}{RGB+Flow} & 58.6 & 54.1 & 45.5 & 32.5 & 17.2 & 41.6 & & 43.6 & 28.7 & 3.7 & 27.8 \\
& \multicolumn{1}{c|}{RTD-Net~\cite{tan2021relaxed}} & \multicolumn{1}{c|}{RGB+Flow} & 68.3 & 62.3 & 51.9 & 38.8 & 23.7 & 49.0 & & 47.2 & 30.7 & 8.6 & 30.8\\
& \multicolumn{1}{c|}{AFSD~\cite{lin2021learning}} & \multicolumn{1}{c|}{RGB+Flow} & 67.3 & 62.4 & 55.5 & 43.7 & 31.1 & 52.0 &  & 52.4 & 35.3 & 6.5 & 34.4\\ 
& \multicolumn{1}{c|}{Aformer~\cite{zhang2022actionformer}} & \multicolumn{1}{c|}{RGB+Flow} & {\bf 82.1} & {\bf 77.8} & {\bf 71.0} & {\bf 59.4} & {\bf 43.9} & {\bf 66.8} & & {\bf 53.5} & {\bf 36.2} & 8.2 & {\bf 35.6} \\
\midrule
\multirow{6}{*}{\begin{tabular}[c]{@{}c@{}}75\% Seen\\ 25\% Unseen\end{tabular}} &
\multicolumn{1}{c|}{I3D~\cite{nag2022zero}} & \multicolumn{1}{c|}{RGB} & 28.5 & 20.3 & 17.1 & 10.5 & 6.9 & 16.6 & & 32.6 & 18.5 & 5.8 & 19.6 \\
& \multicolumn{1}{c|}{CLIP~\cite{Radford21}} & \multicolumn{1}{c|}{RGB} & 33.0 & 25.5 & 18.3 & 11.6 & 5.7 & 18.8 & & 35.6 & 20.4 & 2.1 & 20.2 \\
& \multicolumn{1}{c|}{E-Prompt~\cite{ju2022prompting}} &  \multicolumn{1}{c|}{RGB} & 39.7 & 31.6 & 23.0 & 14.9 & 7.5 & 23.3 &  & 37.6 & 22.9 & 3.8 & 23.1 \\
& \multicolumn{1}{c|}{STALE~\cite{nag2022zero}} & \multicolumn{1}{c|}{RGB} & 40.5 & 32.3 & 23.5 & 15.3 & 7.6 & 23.8 &  & 38.2 & 25.2 & {\bf 6.0} & 24.9 \\ \cline{2-9}  \cline{11-14}
& \multirow{2}{*}{\textbf{Ours}} & \multicolumn{1}{c|}{RGB} & 46.3 & 39.0 & 29.5 & 18.3 & 8.7 & 28.4 &  & 42.0 & 25.8 & 3.2 & 25.9 \\
&  & \multicolumn{1}{c|}{RGB+Flow} & {\bf 64.3} & {\bf 56.9} & {\bf 46.0} & {\bf 32.6} & {\bf 18.5} & {\bf 43.7} & & {\bf 43.9} & {\bf 27.3}  & 3.8 & {\bf 27.5}  \\ 
\midrule 
 \multirow{6}{*}{\begin{tabular}[c]{@{}c@{}}50\% Seen\\ 50\% Unseen\end{tabular}} &
 \multicolumn{1}{c|}{I3D~\cite{nag2022zero}} & \multicolumn{1}{c|}{RGB} & 21.0 & 16.4 & 11.2 & 6.3 & 3.2 & 11.6 & & 25.3 & 13.0 & 3.7 & 12.9 \\
 & \multicolumn{1}{c|}{CLIP~\cite{Radford21}} & \multicolumn{1}{c|}{RGB} & 27.2 & 21.3 & 15.3 & 9.7 & 4.8 & 15.7 & & 28.0 & 16.4 & 1.2 & 16.0 \\
 & \multicolumn{1}{c|}{E-Prompt~\cite{ju2022prompting}} & \multicolumn{1}{c|}{RGB} & 37.2 & 29.6 & 21.6 & 14.0 & 7.2 & 21.9 &  & 32.0 & 19.3 & 2.9 & 19.6 \\
 & \multicolumn{1}{c|}{STALE~\cite{nag2022zero}} & \multicolumn{1}{c|}{RGB} & 38.3 & 30.7 & 21.2 & 13.8 & 7.0 & 22.2 &  & 32.1 & 20.7 & {\bf 5.9} & 20.5 \\  \cline{2-9}  \cline{11-14}
 & \multirow{2}{*}{\textbf{Ours}}
 & \multicolumn{1}{c|}{RGB} & 42.3 & 34.7 & 25.8 & 16.2 & 7.5 & 25.3 & & 34.3 & 20.8 & 3.0 & 21.0 \\
 &  & \multicolumn{1}{c|}{RGB+Flow} & {\bf 55.7} & {\bf 48.9} & {\bf 39.6} & {\bf 28.5} & {\bf 16.1} & {\bf 37.8} & & {\bf 36.5} & {\bf 22.3}  & 3.5 & {\bf 22.5} \\
\bottomrule
\end{tabular}
\caption{\textbf{Comparison with state-of-the-art methods on zero-shot scenarios.} AVG is the average mAP in [0.3:0.1:0.7] on THUMOS14, and [0.5:0.05:0.95] on ActivityNet1.3. Using only RGB, our method has outperformed all zero-shot studies by a large margin. By adding Flow, our method is given explicit motion inputs to be comparable with early closed-set methods.} 
\label{tab:zeroshot}
\end{table*}

In general, making alignment with Text gives RGB or Flow the open-vocabulary ability, enabling them to deal with zero-shot classification. Comparing to adopting RGB only, introducing Flow for tri-modal alignment leads to impressive boosts in the performance of both proposal and classification. For example, 8.5\% gains on oracle average mAP and 3.8\% gains on TOP1 accuracy over only I3D RGB pre-training, revealing the essence of Flow. In terms of efficacy, RGB has more advantages on zero-shot classification, mainly because it gives valuable appearance or context to distinguish actions; while Flow is better at category-agnostic localization, consistent with the closed-set scenarios, {\em i.e.}, explicit motion inputs are critical clues for action discovery.

\vspace{0.2cm}
\noindent \textbf{Encoder generalization.} 
\hspace{1pt} Table~\ref{tab:ablatalign} also attempts to align two types of the RGB encoders with the Flow and Text encoders. As is evident, our method shows promising performance boosts with these two RGB encoders, validating the strong encoder generalization, {\em i.e.}, the same technique should be applicable to more pre-trained encoders. As far as RGB encoder is concerned, the I3D and CLIP have their own advantages. As CLIP is pre-trained using 400M image-text pairs, it shows better vision-language alignment, leading to strong zero-shot classification. While I3D is pre-trained with Kinetics-400 for better temporal continuity, resulting in superior category-agnostic detection, {\em i.e.}, Oracle mAP.

\vspace{-0.3cm}
\subsection{Detailed Comparison \& Module Choice}    \label{subsec:choices}
\vspace{-0.05cm}
In this section, we make detailed comparisons to further dissect model architectures and optimization designs. Experiments are conducted under 75:25 zero-shot splits on THUMOS14, using the I3D encoders for both RGB and Flow, while the CLIP encoder for Text.

\vspace{0.2cm}
\noindent \textbf{Prompt positions.} 
\hspace{1pt} For the learnable prompt vectors $\mathbf{M}_{\mathrm{cond}}$ from vision-conditional prompt module, they can be fed to the input or output of the text encoder, serving as some visual contexts functionally. We make comparisons of these two usage positions (denoting as input and output respectively) in Table~\ref{tab:position}, and observe similar performance. In general, these two positions are almost equivalent as they both extract fine-grained action information from the visual stream to textual stream.

\begin{table*}[t]
\footnotesize
\setlength\tabcolsep{4.2pt}
\centering
\begin{tabular}{c|c|c|ccccccccccccc}   
\toprule 
\multirow{2}{*}{} & \multirow{2}{*}{} & \multirow{2}{*}{} & \multicolumn{7}{c}{THUMOS14~(mAP@IoU)} &  & \multicolumn{5}{c}{ActivityNet1.3~(mAP@IoU)} \\ \cmidrule{4-10} \cmidrule{12-16}
Setting & Method & Modality & \multicolumn{1}{c|}{Shot} & 0.3 & 0.4 & 0.5 & 0.6 & 0.7 & AVG &  & \multicolumn{1}{c|}{Shot} & 0.5 & 0.75 & 0.95 & AVG \\ \midrule
\multirow{6}{*}{\begin{tabular}[c]{@{}c@{}}75\% Seen\\ 25\% Unseen\end{tabular}} & 
\multicolumn{1}{c|}{E-Prompt} & \multicolumn{1}{c|}{RGB} & \multicolumn{1}{c|}{\multirow{3}{*}{1}} & 44.6 & 36.2 & 27.6 & 17.7 & 8.2 & 26.8 &  & \multicolumn{1}{c|}{\multirow{3}{*}{5}} & 39.8 & 23.2 & 1.5 & 23.2 \\
& \multicolumn{1}{c|}{Ours} & \multicolumn{1}{c|}{RGB} & \multicolumn{1}{c|}{} & 46.8 & 39.5 & 29.8 & 18.6 & 8.9 & 28.7 & & \multicolumn{1}{c|}{} & 43.3 & 26.6 & 3.4 & 26.6 \\ 
& \multicolumn{1}{c|}{Ours} & \multicolumn{1}{c|}{RGB+Flow} & \multicolumn{1}{c|}{} & {\bf 64.6} & {\bf 57.0} & {\bf 46.2} & {\bf 33.1} & {\bf 18.6} & {\bf 43.9} &  & \multicolumn{1}{c|}{} & {\bf 45.9} & {\bf 28.3}  & {\bf 4.0} & {\bf 28.5} \\  \cline{2-16} 
& \multicolumn{1}{c|}{E-Prompt} & \multicolumn{1}{c|}{RGB} & \multicolumn{1}{c|}{\multirow{3}{*}{2}} & 45.5 & 36.9 & 28.2 & 18.0 & 8.3 & 27.4 &  & \multicolumn{1}{c|}{\multirow{3}{*}{10}} & 41.2 & 23.8 & 1.5 & 23.9 \\
& \multicolumn{1}{c|}{Ours} & \multicolumn{1}{c|}{RGB} & \multicolumn{1}{c|}{} & 47.6 & 40.1 & 30.1 & 19.1 & 9.1 & 29.2 &  & \multicolumn{1}{c|}{} & 44.2 & 27.1 & 3.6 & 27.1 \\
& \multicolumn{1}{c|}{Ours} & \multicolumn{1}{c|}{RGB+Flow} & \multicolumn{1}{c|}{} & {\bf 65.6} & {\bf 57.8} & {\bf 47.1} & {\bf 33.9} & {\bf 19.0} & {\bf 44.6} &  & \multicolumn{1}{c|}{} & {\bf 46.5} & {\bf 28.5}  & {\bf 4.2} & {\bf 28.8} \\
\midrule
\multirow{6}{*}{\begin{tabular}[c]{@{}c@{}}50\% Seen\\ 50\% Unseen\end{tabular}} 
& \multicolumn{1}{c|}{E-Prompt} & \multicolumn{1}{c|}{RGB} & \multicolumn{1}{c|}{\multirow{3}{*}{1}} & 41.2 & 33.1 & 24.1 & 14.6 & 6.4 & 23.9 &  & \multicolumn{1}{c|}{\multirow{3}{*}{5}}  & 33.6 & 20.8 & 0.5 & 20.4 \\
& \multicolumn{1}{c|}{Ours} & \multicolumn{1}{c|}{RGB} & \multicolumn{1}{c|}{} & 42.6 & 34.9 & 25.9 & 16.3 & 7.6 & 25.5 &  & \multicolumn{1}{c|}{} & 37.3 & 23.1 & 3.2 & 23.0 \\
& \multicolumn{1}{c|}{Ours} & \multicolumn{1}{c|}{RGB+Flow} & \multicolumn{1}{c|}{} & {\bf 57.3} & {\bf 50.7} & {\bf 41.4} & {\bf 29.7} & {\bf 16.7} & {\bf 39.2} &  & \multicolumn{1}{c|}{} & {\bf 39.2} & {\bf 24.3} & {\bf 3.7} & {\bf 24.5} \\ \cline{2-16}
& \multicolumn{1}{c|}{E-Prompt} & \multicolumn{1}{c|}{RGB} & \multicolumn{1}{c|}{\multirow{3}{*}{2}} & 42.4 & 34.2 & 25.0 & 15.2 & 6.8 & 24.7 &  & \multicolumn{1}{c|}{\multirow{3}{*}{10}} & 34.9 & 21.1 & 0.6 & 21.0 \\
& \multicolumn{1}{c|}{Ours} & \multicolumn{1}{c|}{RGB} & \multicolumn{1}{c|}{} & 43.6 & 35.7 & 26.6 & 16.7 & 7.7 & 26.0 &  & \multicolumn{1}{c|}{} & 38.2 & 23.5 & 3.4 & 23.3 \\
& \multicolumn{1}{c|}{Ours} & \multicolumn{1}{c|}{RGB+Flow} & \multicolumn{1}{c|}{} & {\bf 60.1} & {\bf 53.2} & {\bf 43.5} & {\bf 31.2} & {\bf 17.7} & {\bf 41.2} &  & \multicolumn{1}{c|}{} & {\bf 40.3} & {\bf 24.9} & {\bf 3.8} & {\bf 25.1} \\
\bottomrule
\end{tabular}
\caption{\textbf{Comparison with state-of-the-art methods on few-shot scenarios.} We retrain and report few-shot results of E-Prompt with their released codes. Although only several support samples are given for novel categories, few-shot scenarios obtain considerable gains over zero-shot scenarios. Flow can further boost the RGB performance on both datasets.}
\label{tab:fewshot}
\end{table*}

\vspace{0.2cm}
\noindent \textbf{Prompt length \& format.} 
\hspace{1pt} For the learned prompt vectors $\mathbf{M}_{\mathrm{cond}}$, we prepend or append them in the format of $[\mathbf{M}_{\mathrm{cond}},\, \Phi_\textsc{tokenise}(\mathcal{C}_{\mathrm{name}}),\, \mathbf{M}_{\mathrm{cond}}]$. In practise, this format is equivalent with $[\mathbf{M}_{\mathrm{cond}},\, \Phi_\textsc{tokenize}(\mathcal{C}_{\mathrm{name}})]$ or $[\Phi_\textsc{tokenize}(\mathcal{C}_{\mathrm{name}}),\, \mathbf{M}_{\mathrm{cond}}]$, as revealed by existing studies~\cite{ju2022prompting,zhou2022conditional}. On the other hand, given $\mathbf{M}_{\mathrm{cond}} \in \mathbb{R}^{K \times D_{\mathrm{text}}}$, to validate the effect of prompt length $K$, we also experiment in Figure~\ref{fig:length}, {\em i.e.}, gradually increase the prompt number from $4$ to $64$. Overall, the variance in prompt length has marginal effects on performance. We therefore pick $32$ prompt vectors for its good trade-off between model performance and parameter efficiency.

\vspace{0.2cm}
\noindent \textbf{Sharing visual backbone.} 
\hspace{1pt} In the closed-set action localization, there are two public backbone strategies: share one network (early fusion) or utilize separate networks (late fusion) for RGB and Flow. In Table~\ref{tab:share}, we explore these two strategies for low-shot scenarios. Generally speaking, sharing a visual backbone damages the performance to some extent, both for proposal and classification. This is possibly due to the premature fusion of RGB and Flow, hindering the effective alignment between these two visual modalities.

\vspace{0.2cm}
\noindent \textbf{Freeze encoders {\em vs.} end-to-end fine-tuning.} 
\hspace{1pt} To avoid heavy computational burdens, we freeze the pre-trained encoders for RGB-Flow-Text modalities, but only optimize lightweight modules (detector, regressor, etc.) in our proposed method. Here, in Table~\ref{tab:fixend}, we compare with end-to-end fine-tuning of the entire model on THUMOS14. Surprisingly, fine-tuning more parameters actually leads to lower performance. We conjecture this is because the large models can be overfitting to the training data, thus damaging the generalization towards unseen (novel) action categories.

\vspace{-0.2cm}
\subsection{Comparison with state-of-the-art methods}  \label{subsec:SOTA}
\vspace{-0.1cm}
This section makes full comparisons with state-of-the-art methods on both THUMOS14 and ActivityNet1.3. For the sake of fairness, we employ the CLIP encoders for both RGB and Text modalities. While for the Flow modality, we adopt the I3D network pre-trained on the Kinetics-400 dataset~\cite{carreira2017quo}. For low-shot classifiers, we here use vision-conditional prompt tuning ($\mathbf{M}_{\mathrm{cond}}$).

\begin{figure}[t]
\begin{center}
\vspace{0.1cm}
\includegraphics[width=0.4\textwidth] {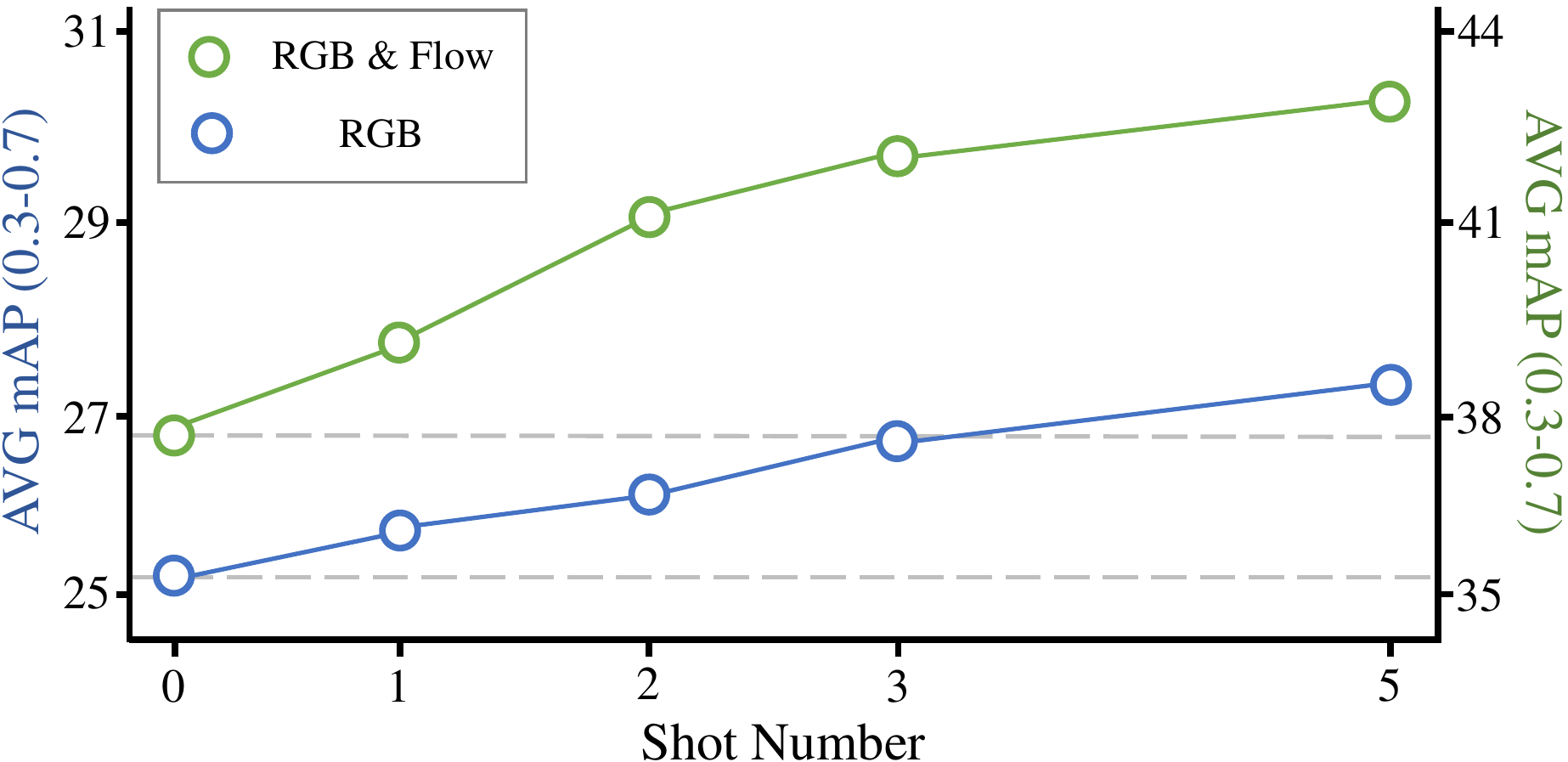}
\end{center}
\vspace{-0.4cm}
\caption{\textbf{Effect of shot number.} More shots for novel categories bring greater gains, but also heavier annotation costs.}
\label{fig:shot}
\end{figure}

\vspace{0.2cm}
\noindent \textbf{The zero-shot performance} is reported in Table~\ref{tab:zeroshot}. We respectively list the results of RGB and Flow for clear understanding. In general, on all benchmarks, our framework achieves new state-of-the-art under most IoU regimes, using the single RGB modality, for example, comparing to concurrent work, we significantly surpass competitive methods by over 5\% average mAP on THUMOS14. Overall, existing zero-shot TAL methods usually adopt prompt tuning in the text stream, lacking the understanding of visual details. Instead, our vision-conditional method uses RGB or Flow embeddings to enrich the classifier generation, showing better generalization to the novel action categories.

Moreover, adding optical Flows for explicit motion inputs brings immediate improvements, proving the effectiveness of tri-modal alignment. Delightfully, our zero-shot results with RGB-Flow inputs are even comparable with several early methods~\cite{Chao18,Lin18,shou2017cdc} from closed-set scenarios, demonstrating the superiority. Additionally, comparing to the 75:25 data splits, the 50:50 data splits pose bigger challenges, yet our method still yields excellent results, showing powerful generalization.

\begin{figure*}[t]
\begin{center}
\includegraphics[width=0.98\textwidth] {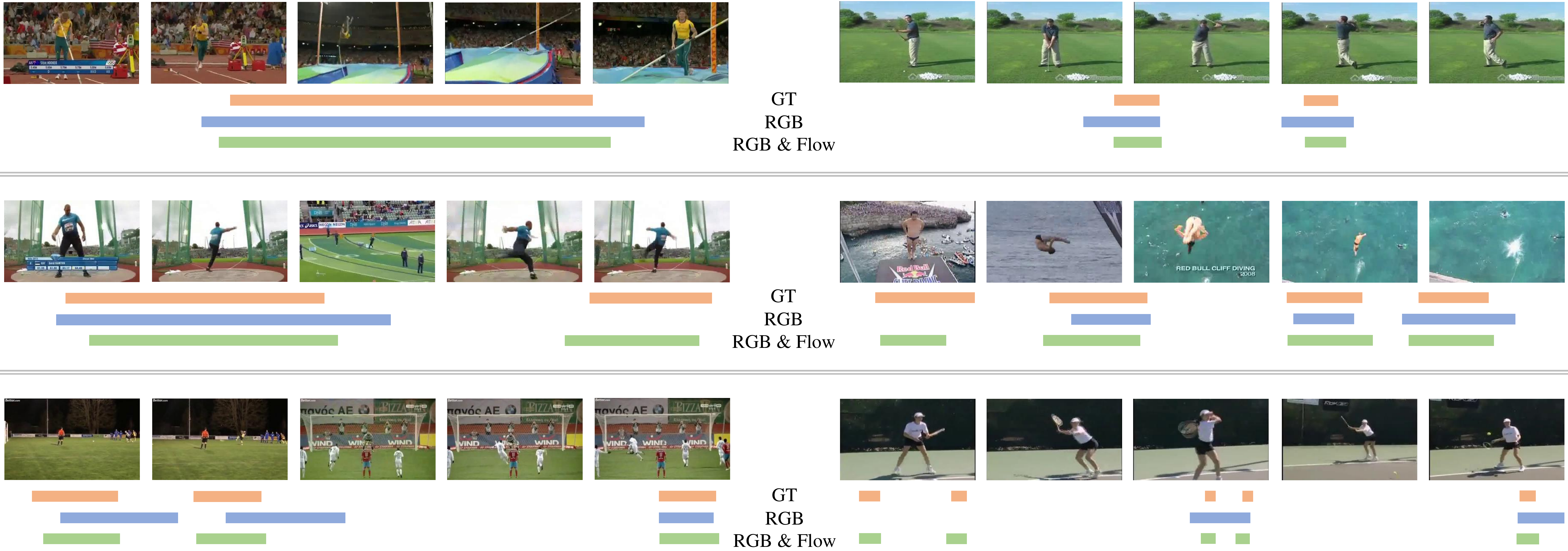}
\end{center}
\vspace{-0.2cm}
\caption{\textbf{Qualitative zero-shot results on THUMOS14.} For various videos from novel categories, our method all outputs good detection results, although action number and action duration vary frequently. Single RGB sometimes has large deviations, or even omits action instances. By bringing motion details, Flow could further correct or complete the RGB results.}
\vspace{0.2cm}
\label{fig:results}
\end{figure*}

{\bf Note that}, comparing to THUMOS14, ActivityNet has lower requirements for classification and detection, as there are only $1.5$ action instances per video, and most videos contain only one category; while for THUMOS14, there are an average of $15$ action instances per video, with significant variations. This phenomenon has become a consensus~\cite{Chao18,zhang2019adversarial,luo2020weakly} of temporal localization, and could somewhat limit our gains on ActivityNet.

\vspace{0.2cm}
\noindent \textbf{Few-shot performance.}
\hspace{1pt} For few-shot scenarios, several video exemplars are annotated for novel (unseen) categories as the support set. Since we are not aware of any existing benchmarks, we initiate evaluation settings based on data splits of zero-shot scenarios. Specifically, we label $1$ or $2$ videos (shots) each category for THUMOS14, while label $5$ or $10$ videos per category for ActivityNet1.3. The shot number is set according to the dataset scale. We conduct $3$ trials and report the average results to ensure statistical significance.

As shown in Table~\ref{tab:fewshot}, we also retrain E-Prompt~\cite{ju2022prompting} with its released codes to get few-shot results for comparison. In all settings of both datasets, these few-shot video exemplars bring considerable gains over the zero-shot counterparts, by providing explicit distribution of novel categories. In addition, more shots naturally bring higher performance gains, and optical flows still have immediate improvements in performance.

To further evaluate the efficacy of shot number, Figure~\ref{fig:shot} reports some results under the 50:50 data splits on THUMOS14. As can be seen, more shots for novel categories indeed bring greater gains, and also enhance the model generalization. However, there is a price to pay, {\em i.e.}, annotation overheads continue to rise.

\vspace{0.2cm}
\noindent \textbf{Effectiveness of optical flow.} 
On both datasets, we can observe that adding Flow consistently improves the performance by a large margin, for both zero-shot and few-shot tasks. Especially on THUMOS14, introducing optical Flow boosts the average mAP for more than 10\%. The same phenomenon could be observed under various data splits, thus reflecting the significance and effectiveness of tri-modal alignment.

\vspace{-0.2cm}
\subsection{Qualitative Localization Results}  \label{subsec:qualitative}
\vspace{-0.05cm}
We visualize several detection results of novel categories in Figure~\ref{fig:results}, under the 75:25 zero-shot splits on THUMOS14. Note that one proposal is shown only if the predicted category is correct. As is evident, for various novel categories, both action number and action duration could vary frequently in these videos, posing great challenges to TAL models. Nevertheless, our method obtains good results in most cases, again proving the gratifying effectiveness. Besides, the single RGB modality sometimes outputs large deviations, or even omits action instances. By adding motion details, the Flow modality can further correct or complete results.

\vspace{-0.1cm}
\section{Conclusion}
This paper considers low-shot temporal action localization, and handles one main challenge, {\em i.e.}, the lexical ambiguities in vanilla category names. To enrich the discriminative power of models, we decompose categories into adaptive attribute descriptions, by prompting large-scale language models, making text-based classifiers more detailed. Moreover, to tackle cases where it is difficult to give comprehensive descriptions, we design one novel vision-conditional prompt module, inputting RGB \& Flow embeddings to generate prompting with rich visual details, for powerful vision-based classifiers. Besides, we also inject optical flows for explicit motion inputs, bringing impressive category-agnostic detection. Extensive experiments and thorough ablations demonstrate the effectiveness of core components, and our superior performance over state-of-the-art methods.

\bibliographystyle{spmpsci}     
\bibliography{egbib}

\begin{thebibliography}{10}
\providecommand{\url}[1]{{#1}}
\providecommand{\urlprefix}{URL }
\expandafter\ifx\csname urlstyle\endcsname\relax
  \providecommand{\doi}[1]{DOI~\discretionary{}{}{}#1}\else
  \providecommand{\doi}{DOI~\discretionary{}{}{}\begingroup
  \urlstyle{rm}\Url}\fi

\bibitem{bai2020boundary}
Bai, Y., Wang, Y., Tong, Y., Yang, Y., Liu, Q., Liu, J.: Boundary content graph
  neural network for temporal action proposal generation.
\newblock In: Proceedings of the European Conference on Computer Vision, pp.
  121--137 (2020)

\bibitem{bao2022opental}
Bao, W., Yu, Q., Kong, Y.: Opental: Towards open set temporal action
  localization.
\newblock In: Proceedings of the IEEE Conference on Computer Vision and Pattern
  Recognition, pp. 2979--2989 (2022)

\bibitem{Brown20}
Brown, T., Mann, B., Ryder, N., Subbiah, M., Kaplan, J.D., Dhariwal, P.,
  Neelakantan, A., Shyam, P., Sastry, G., Askell, A., Agarwal, S.,
  Herbert-Voss, A., Krueger, G., Henighan, T., Child, R., Ramesh, A., Ziegler,
  D., Wu, J., Winter, C., Hesse, C., Chen, M., Sigler, E., Litwin, M., Gray,
  S., Chess, B., Clark, J., Berner, C., McCandlish, S., Radford, A., Sutskever,
  I., Amodei, D.: Language models are few-shot learners.
\newblock In: Advances in Neural Information Processing Systems, pp. 1877--1901
  (2020)

\bibitem{caba2015activitynet}
Caba~Heilbron, F., Escorcia, V., Ghanem, B., Carlos~Niebles, J.: Activitynet: A
  large-scale video benchmark for human activity understanding.
\newblock In: Proceedings of the IEEE Conference on Computer Vision and Pattern
  Recognition, pp. 961--970 (2015)

\bibitem{carreira2017quo}
Carreira, J., Zisserman, A.: Quo vadis, action recognition? a new model and the
  kinetics dataset.
\newblock In: Proceedings of the IEEE Conference on Computer Vision and Pattern
  Recognition, pp. 6299--6308 (2017)

\bibitem{Chao18}
Chao, Y.W., Vijayanarasimhan, S., Seybold, B., Ross, D.A., Deng, J.,
  Sukthankar, R.: Rethinking the faster r-cnn architecture for temporal action
  localisation.
\newblock In: Proceedings of the IEEE Conference on Computer Vision and Pattern
  Recognition, pp. 1130--1139 (2018)

\bibitem{chao2018rethinking}
Chao, Y.W., Vijayanarasimhan, S., Seybold, B., Ross, D.A., Deng, J.,
  Sukthankar, R.: Rethinking the faster r-cnn architecture for temporal action
  localization.
\newblock In: Proceedings of the IEEE Conference on Computer Vision and Pattern
  Recognition, pp. 1130--1139 (2018)

\bibitem{dai2022ms}
Dai, R., Das, S., Kahatapitiya, K., Ryoo, M.S., Bremond, F.: Ms-tct:
  Multi-scale temporal convtransformer for action detection.
\newblock In: Proceedings of the IEEE Conference on Computer Vision and Pattern
  Recognition, pp. 20041--20051 (2022)

\bibitem{Frome13}
Frome, A., Corrado, G.S., Shlens, J., Bengio, S., Dean, J., Ranzato, M.A.,
  Mikolov, T.: Devise: A deep visual-semantic embedding model.
\newblock In: Advances in Neural Information Processing Systems (2013)

\bibitem{gao2018ctap}
Gao, J., Chen, K., Nevatia, R.: Ctap: Complementary temporal action proposal
  generation.
\newblock In: Proceedings of the European Conference on Computer Vision, pp.
  68--83 (2018)

\bibitem{gao2022fine}
Gao, J., Chen, M., Xu, C.: Fine-grained temporal contrastive learning for
  weakly-supervised temporal action localization.
\newblock In: Proceedings of the IEEE Conference on Computer Vision and Pattern
  Recognition, pp. 19999--20009 (2022)

\bibitem{gao2017turn}
Gao, J., Yang, Z., Chen, K., Sun, C., Nevatia, R.: Turn tap: Temporal unit
  regression network for temporal action proposals.
\newblock In: Proceedings of the International Conference on Computer Vision,
  pp. 3628--3636 (2017)

\bibitem{Gao21}
Gao, T., Fisch, A., Chen, D.: Making pre-trained language models better
  few-shot learners.
\newblock In: Association for Computational Linguistics (2021)

\bibitem{he2022asm}
He, B., Yang, X., Kang, L., Cheng, Z., Zhou, X., Shrivastava, A.: Asm-loc:
  Action-aware segment modeling for weakly-supervised temporal action
  localization.
\newblock In: Proceedings of the IEEE Conference on Computer Vision and Pattern
  Recognition, pp. 13925--13935 (2022)

\bibitem{huang2021foreground}
Huang, L., Wang, L., Li, H.: Foreground-action consistency network for weakly
  supervised temporal action localization.
\newblock In: Proceedings of the International Conference on Computer Vision,
  pp. 8002--8011 (2021)

\bibitem{Jia21}
Jia, C., Yang, Y., Xia, Y., Chen, Y.T., Parekh, Z., Pham, H., Le, Q.V., Sung,
  Y., Li, Z., Duerig, T.: Scaling up visual and vision-language representation
  learning with noisy text supervision.
\newblock In: Proceedings of the International Conference on Machine Learning,
  pp. 4904--4916 (2021)

\bibitem{jia2022visual}
Jia, M., Tang, L., Chen, B.C., Cardie, C., Belongie, S., Hariharan, B., Lim,
  S.N.: Visual prompt tuning.
\newblock In: Proceedings of the European Conference on Computer Vision, pp.
  709--727 (2022)

\bibitem{jiang2014thumos}
Jiang, Y.G., Liu, J., Zamir, A.R., Toderici, G., Laptev, I., Shah, M.,
  Sukthankar, R.: Thumos challenge: Action recognition with a large number of
  classes.
\newblock \urlprefix\url{{http://crcv.ucf.edu/THUMOS14/}}

\bibitem{Jiang20}
Jiang, Z., Xu, F.F., Araki, J., Neubig, G.: How can we know what language
  models know?
\newblock Transactions of the Association for Computational Linguistics pp.
  423--438 (2020)

\bibitem{ju2022prompting}
Ju, C., Han, T., Zheng, K., Zhang, Y., Xie, W.: Prompting visual-language
  models for efficient video understanding.
\newblock In: Proceedings of the European Conference on Computer Vision, pp.
  105--124. Springer (2022)

\bibitem{ju2023constraint}
Ju, C., Wang, H., Liu, J., Ma, C., Zhao, P., Zhang, Y., Chang, J., Tian, Q.:
  Constraint and union for partially-supervised temporal sentence grounding.
\newblock arXiv preprint arXiv:2302.09850  (2023)

\bibitem{ju2021divide}
Ju, C., Zhao, P., Chen, S., Zhang, Y., Wang, Y., Tian, Q.: Divide and conquer
  for single-frame temporal action localization.
\newblock In: Proceedings of the International Conference on Computer Vision,
  pp. 13455--13464 (2021)

\bibitem{ju2021adaptive}
Ju, C., Zhao, P., Chen, S., Zhang, Y., Zhang, X., Tian, Q.: Adaptive mutual
  supervision for weakly-supervised temporal action localization.
\newblock IEEE Transactions on Multimedia  (2022)

\bibitem{ju2020point}
Ju, C., Zhao, P., Zhang, Y., Wang, Y., Tian, Q.: Point-level temporal action
  localization: Bridging fully-supervised proposals to weakly-supervised
  losses.
\newblock arXiv preprint arXiv:2012.08236  (2020)

\bibitem{ju2022distilling}
Ju, C., Zheng, K., Liu, J., Zhao, P., Zhang, Y., Chang, J., Wang, Y., Tian, Q.:
  Distilling vision-language pre-training to collaborate with weakly-supervised
  temporal action localization.
\newblock arXiv preprint arXiv:2212.09335  (2022)

\bibitem{ke2010volumetric}
Ke, Y., Sukthankar, R., Hebert, M.: Volumetric features for video event
  detection.
\newblock International Journal of Computer Vision \textbf{88}, 339--362 (2010)

\bibitem{lee2021learning}
Lee, P., Byun, H.: Learning action completeness from points for
  weakly-supervised temporal action localization.
\newblock In: Proceedings of the International Conference on Computer Vision,
  pp. 13648--13657 (2021)

\bibitem{lee2019background}
Lee, P., Uh, Y., Byun, H.: Background suppression network for weakly-supervised
  temporal action localization.
\newblock In: Proceedings of the AAAI Conference on Artificial Intelligence,
  pp. 11320--11327 (2020)

\bibitem{lee2021weakly}
Lee, P., Wang, J., Lu, Y., Byun, H.: Weakly-supervised temporal action
  localization by uncertainty modeling.
\newblock In: Proceedings of the AAAI Conference on Artificial Intelligence,
  pp. 1854--1862 (2021)

\bibitem{Lester21}
Lester, B., Al-Rfou, R., Constant, N.: The power of scale for
  parameter-efficient prompt tuning.
\newblock In: Proceedings of the Conference on Empirical Methods in Natural
  Language Processinng (2021)

\bibitem{li21-prefixtuning}
Li, X.L., Liang, P.: Prefix-tuning: Optimizing continuous prompts for
  generation.
\newblock In: Association for Computational Linguistics (2021)

\bibitem{li2021prefix}
Li, X.L., Liang, P.: Prefix-tuning: Optimizing continuous prompts for
  generation.
\newblock In: Association for Computational Linguistics (2021)

\bibitem{lin2021learning}
Lin, C., Xu, C., Luo, D., Wang, Y., Tai, Y., Wang, C., Li, J., Huang, F., Fu,
  Y.: Learning salient boundary feature for anchor-free temporal action
  localization.
\newblock In: Proceedings of the IEEE Conference on Computer Vision and Pattern
  Recognition, pp. 3320--3329 (2021)

\bibitem{lin2019bmn}
Lin, T., Liu, X., Li, X., Ding, E., Wen, S.: Bmn: Boundary-matching network for
  temporal action proposal generation.
\newblock In: Proceedings of the International Conference on Computer Vision,
  pp. 3889--3898 (2019)

\bibitem{lin2017single}
Lin, T., Zhao, X., Shou, Z.: Single shot temporal action detection.
\newblock In: Proceedings of ACM International Conference on Multimedia, pp.
  988--996 (2017)

\bibitem{lin2018bsn}
Lin, T., Zhao, X., Su, H., Wang, C., Yang, M.: Bsn: Boundary sensitive network
  for temporal action proposal generation.
\newblock In: Proceedings of the European Conference on Computer Vision, pp.
  3--19 (2018)

\bibitem{Lin18}
Lin, T., Zhao, X., Su, H., Wang, C., Yang, M.: {BSN}: Boundary sensitive
  network for temporal action proposal generation.
\newblock In: Proceedings of the European Conference on Computer Vision, pp.
  3--19 (2018)

\bibitem{liu2019completeness}
Liu, D., Jiang, T., Wang, Y.: Completeness modeling and context separation for
  weakly supervised temporal action localization.
\newblock In: Proceedings of the IEEE Conference on Computer Vision and Pattern
  Recognition, pp. 1298--1307 (2019)

\bibitem{liu2019multi}
Liu, Y., Ma, L., Zhang, Y., Liu, W., Chang, S.F.: Multi-granularity generator
  for temporal action proposal.
\newblock In: Proceedings of the IEEE Conference on Computer Vision and Pattern
  Recognition, pp. 3604--3613 (2019)

\bibitem{liu2019weakly}
Liu, Z., Wang, L., Zhang, Q., Gao, Z., Niu, Z., Zheng, N., Hua, G.: Weakly
  supervised temporal action localization through contrast based evaluation
  networks.
\newblock In: Proceedings of the International Conference on Computer Vision,
  pp. 3899--3908 (2019)

\bibitem{lu2022prompt}
Lu, Y., Liu, J., Zhang, Y., Liu, Y., Tian, X.: Prompt distribution learning.
\newblock In: Proceedings of the IEEE Conference on Computer Vision and Pattern
  Recognition, pp. 5206--5215 (2022)

\bibitem{luo2022clip4clip}
Luo, H., Ji, L., Zhong, M., Chen, Y., Lei, W., Duan, N., Li, T.: Clip4clip: An
  empirical study of clip for end to end video clip retrieval and captioning.
\newblock Neurocomputing pp. 293--304 (2022)

\bibitem{luo2021action}
Luo, W., Zhang, T., Yang, W., Liu, J., Mei, T., Wu, F., Zhang, Y.: Action unit
  memory network for weakly supervised temporal action localization.
\newblock In: Proceedings of the IEEE Conference on Computer Vision and Pattern
  Recognition, pp. 9969--9979 (2021)

\bibitem{luo2020weakly}
Luo, Z., Guillory, D., Shi, B., Ke, W., Wan, F., Darrell, T., Xu, H.:
  Weakly-supervised action localization with expectation-maximization
  multi-instance learning.
\newblock In: Proceedings of the European Conference on Computer Vision, pp.
  729--745 (2020)

\bibitem{ma2023diffusionseg}
Ma, C., Yang, Y., Ju, C., Zhang, F., Liu, J., Wang, Y., Zhang, Y., Wang, Y.:
  Diffusionseg: Adapting diffusion towards unsupervised object discovery.
\newblock arXiv preprint arXiv:2303.09813  (2023)

\bibitem{ma2022weakly}
Ma, F., Zhu, L., Yang, Y.: Weakly supervised moment localization with decoupled
  consistent concept prediction.
\newblock International Journal of Computer Vision \textbf{130}(5), 1244--1258
  (2022)

\bibitem{ma2020sf}
Ma, F., Zhu, L., Yang, Y., Zha, S., Kundu, G., Feiszli, M., Shou, Z.: Sf-net:
  Single-frame supervision for temporal action localization.
\newblock In: Proceedings of the European Conference on Computer Vision, pp.
  420--437. Springer (2020)

\bibitem{mettes2019pointly}
Mettes, P., Snoek, C.G.: Pointly-supervised action localization.
\newblock International Journal of Computer Vision \textbf{127}, 263--281
  (2019)

\bibitem{min2020adversarial}
Min, K., Corso, J.J.: Adversarial background-aware loss for weakly-supervised
  temporal activity localization.
\newblock In: Proceedings of the European Conference on Computer Vision, pp.
  283--299 (2020)

\bibitem{Mori99}
Mori, Y., Takahashi, H., Oka, R.: Image-to-word transformation based on
  dividing and vector quantizing images with words.
\newblock In: First International Workshop on Multimedia Intelligent Storage
  and Retrieval Management (ACM Multimedia Conference), pp. 1--9 (1999)

\bibitem{nag2022zero}
Nag, S., Zhu, X., Song, Y.Z., Xiang, T.: Zero-shot temporal action detection
  via vision-language prompting.
\newblock In: Proceedings of the European Conference on Computer Vision, pp.
  681--697. Springer (2022)

\bibitem{narayan2021d2}
Narayan, S., Cholakkal, H., Hayat, M., Khan, F.S., Yang, M.H., Shao, L.:
  D2-net: Weakly-supervised action localization via discriminative embeddings
  and denoised activations.
\newblock In: Proceedings of the International Conference on Computer Vision,
  pp. 13608--13617 (2021)

\bibitem{narayan20193c}
Narayan, S., Cholakkal, H., Khan, F.S., Shao, L.: 3c-net: Category count and
  center loss for weakly-supervised action localization.
\newblock In: Proceedings of the International Conference on Computer Vision,
  pp. 8679--8687 (2019)

\bibitem{nguyen2018weakly}
Nguyen, P., Liu, T., Prasad, G., Han, B.: Weakly supervised action localization
  by sparse temporal pooling network.
\newblock In: Proceedings of the IEEE Conference on Computer Vision and Pattern
  Recognition, pp. 6752--6761 (2018)

\bibitem{nguyen2019weakly}
Nguyen, P.X., Ramanan, D., Fowlkes, C.C.: Weakly-supervised action localization
  with background modeling.
\newblock In: Proceedings of the International Conference on Computer Vision,
  pp. 5502--5511 (2019)

\bibitem{ni2022expanding}
Ni, B., Peng, H., Chen, M., Zhang, S., Meng, G., Fu, J., Xiang, S., Ling, H.:
  Expanding language-image pretrained models for general video recognition.
\newblock In: Proceedings of the European Conference on Computer Vision, pp.
  1--18 (2022)

\bibitem{paul2018w}
Paul, S., Roy, S., Roy-Chowdhury, A.: W-talc: Weakly-supervised temporal
  activity localization and classification.
\newblock In: Proceedings of the European Conference on Computer Vision, pp.
  563--579 (2018)

\bibitem{qing2021temporal}
Qing, Z., Su, H., Gan, W., Wang, D., Wu, W., Wang, X., Qiao, Y., Yan, J., Gao,
  C., Sang, N.: Temporal context aggregation network for temporal action
  proposal refinement.
\newblock In: Proceedings of the International Conference on Computer Vision,
  pp. 485--494 (2021)

\bibitem{Radford21}
Radford, A., Kim, J.W., Hallacy, C., Ramesh, A., Goh, G., Agarwal, S., Sastry,
  G., Askell, A., Mishkin, P., Clark, J., Krueger, G., Sutskever, I.: Learning
  transferable visual models from natural language supervision.
\newblock In: Proceedings of the International Conference on Machine Learning,
  pp. 8748--8763 (2021)

\bibitem{Timo21}
Schick, T., Schütze, H.: Exploiting cloze questions for few shot text
  classification and natural language inference.
\newblock In: In Proceedings of the 16th Conference of the European Chapter of
  the Association for Computer Linguistics (2021)

\bibitem{shi2020weakly}
Shi, B., Dai, Q., Mu, Y., Wang, J.: Weakly-supervised action localization by
  generative attention modeling.
\newblock In: Proceedings of the IEEE Conference on Computer Vision and Pattern
  Recognition, pp. 1009--1019 (2020)

\bibitem{Shin20}
Shin, T., Razeghi, Y., IV, R.L.L., Wallace, E., Singh, S.: {AutoPrompt}:
  Eliciting knowledge from language models with automatically generated
  prompts.
\newblock In: Proceedings of the Conference on Empirical Methods in Natural
  Language Processinng (2020)

\bibitem{shou2017cdc}
Shou, Z., Chan, J., Zareian, A., Miyazawa, K., Chang, S.F.: Cdc:
  Convolutional-de-convolutional networks for precise temporal action
  localization in untrimmed videos.
\newblock In: Proceedings of the IEEE Conference on Computer Vision and Pattern
  Recognition, pp. 5734--5743 (2017)

\bibitem{shou2018autoloc}
Shou, Z., Gao, H., Zhang, L., Miyazawa, K., Chang, S.F.: Autoloc:
  Weakly-supervised temporal action localization in untrimmed videos.
\newblock In: Proceedings of the European Conference on Computer Vision, pp.
  154--171 (2018)

\bibitem{shou2016temporal}
Shou, Z., Wang, D., Chang, S.F.: Temporal action localization in untrimmed
  videos via multi-stage cnns.
\newblock In: Proceedings of the IEEE Conference on Computer Vision and Pattern
  Recognition, pp. 1049--1058 (2016)

\bibitem{shu2015joint}
Shu, T., Xie, D., Rothrock, B., Todorovic, S., Chun~Zhu, S.: Joint inference of
  groups, events and human roles in aerial videos.
\newblock In: CVPR, pp. 4576--4584 (2015)

\bibitem{tan2021relaxed}
Tan, J., Tang, J., Wang, L., Wu, G.: Relaxed transformer decoders for direct
  action proposal generation.
\newblock In: Proceedings of the International Conference on Computer Vision,
  pp. 13526--13535 (2021)

\bibitem{vo2023aoe}
Vo, K., Truong, S., Yamazaki, K., Raj, B., Tran, M.T., Le, N.: Aoe-net:
  Entities interactions modeling with adaptive attention mechanism for temporal
  action proposals generation.
\newblock International Journal of Computer Vision \textbf{131}(1), 302--323
  (2023)

\bibitem{wang2017untrimmednets}
Wang, L., Xiong, Y., Lin, D., Van~Gool, L.: Untrimmednets for weakly supervised
  action recognition and detection.
\newblock In: Proceedings of the IEEE Conference on Computer Vision and Pattern
  Recognition, pp. 4325--4334 (2017)

\bibitem{wang2022rcl}
Wang, Q., Zhang, Y., Zheng, Y., Pan, P.: Rcl: Recurrent continuous localization
  for temporal action detection.
\newblock In: Proceedings of the IEEE Conference on Computer Vision and Pattern
  Recognition, pp. 13566--13575 (2022)

\bibitem{wedel2009improved}
Wedel, A., Pock, T., Zach, C., Bischof, H., Cremers, D.: An improved algorithm
  for tv-l 1 optical flow.
\newblock In: Statistical and geometrical approaches to visual motion analysis,
  pp. 23--45. Springer (2009)

\bibitem{Weston11}
Weston, J., Bengio, S., Usunier, N.: {WSABIE}: Scaling up to large vocabulary
  image annotation.
\newblock In: Proceedings of the International Joint Conference on Artificial
  Intelligence (2011)

\bibitem{xu2017r}
Xu, H., Das, A., Saenko, K.: R-c3d: Region convolutional 3d network for
  temporal activity detection.
\newblock In: Proceedings of the International Conference on Computer Vision,
  pp. 5783--5792 (2017)

\bibitem{xu2021videoclip}
Xu, H., Ghosh, G., Huang, P.Y., Okhonko, D., Aghajanyan, A., Metze, F.,
  Zettlemoyer, L., Feichtenhofer, C.: Videoclip: Contrastive pre-training for
  zero-shot video-text understanding.
\newblock arXiv preprint arXiv:2109.14084  (2021)

\bibitem{xu2020g}
Xu, M., Zhao, C., Rojas, D.S., Thabet, A., Ghanem, B.: G-tad: Sub-graph
  localization for temporal action detection.
\newblock In: Proceedings of the IEEE Conference on Computer Vision and Pattern
  Recognition, pp. 10156--10165 (2020)

\bibitem{xu2019segregated}
Xu, Y., Zhang, C., Cheng, Z., Xie, J., Niu, Y., Pu, S., Wu, F.: Segregated
  temporal assembly recurrent networks for weakly supervised multiple action
  detection.
\newblock In: Proceedings of the AAAI Conference on Artificial Intelligence,
  pp. 9070--9078 (2019)

\bibitem{yang2021background}
Yang, L., Han, J., Zhao, T., Lin, T., Zhang, D., Chen, J.: Background-click
  supervision for temporal action localization.
\newblock IEEE Transactions on Pattern Analysis and Machine Intelligence pp.
  9814--9829 (2021)

\bibitem{yang2020revisiting}
Yang, L., Peng, H., Zhang, D., Fu, J., Han, J.: Revisiting anchor mechanisms
  for temporal action localization.
\newblock IEEE Transactions on Image Processing pp. 8535--8548 (2020)

\bibitem{yang2022learning}
Yang, X., Zhang, H., Gao, C., Cai, J.: Learning to collocate visual-linguistic
  neural modules for image captioning.
\newblock International Journal of Computer Vision pp. 1--19 (2022)

\bibitem{yao2021filip}
Yao, L., Huang, R., Hou, L., Lu, G., Niu, M., Xu, H., Liang, X., Li, Z., Jiang,
  X., Xu, C.: Filip: Fine-grained interactive language-image pre-training.
\newblock In: Proceedings of the International Conference on Learning
  Representations (2022)

\bibitem{yuan2021florence}
Yuan, L., Chen, D., Chen, Y.L., Codella, N., Dai, X., Gao, J., Hu, H., Huang,
  X., Li, B., Li, C., et~al.: Florence: A new foundation model for computer
  vision.
\newblock arXiv preprint arXiv:2111.11432  (2021)

\bibitem{yudistira2022weakly}
Yudistira, N., Kavitha, M.S., Kurita, T.: Weakly-supervised action
  localization, and action recognition using global--local attention of 3d cnn.
\newblock International Journal of Computer Vision \textbf{130}(10), 2349--2363
  (2022)

\bibitem{zeng2019graph}
Zeng, R., Huang, W., Tan, M., Rong, Y., Zhao, P., Huang, J., Gan, C.: Graph
  convolutional networks for temporal action localization.
\newblock In: Proceedings of the International Conference on Computer Vision,
  pp. 7094--7103 (2019)

\bibitem{zhai2022lit}
Zhai, X., Wang, X., Mustafa, B., Steiner, A., Keysers, D., Kolesnikov, A.,
  Beyer, L.: Lit: Zero-shot transfer with locked-image text tuning.
\newblock In: Proceedings of the IEEE Conference on Computer Vision and Pattern
  Recognition, pp. 18123--18133 (2022)

\bibitem{zhang2022actionformer}
Zhang, C., Wu, J., Li, Y.: Actionformer: Localizing moments of actions with
  transformers.
\newblock In: Proceedings of the European Conference on Computer Vision, pp.
  492--510. Springer (2022)

\bibitem{zhang2019adversarial}
Zhang, C., Xu, Y., Cheng, Z., Niu, Y., Pu, S., Wu, F., Zou, F.: Adversarial
  seeded sequence growing for weakly-supervised temporal action localization.
\newblock In: Proceedings of ACM International Conference on Multimedia, pp.
  738--746 (2019)

\bibitem{zhang2022opt}
Zhang, S., Roller, S., Goyal, N., Artetxe, M., Chen, M., Chen, S., Dewan, C.,
  Diab, M., Li, X., Lin, X.V., et~al.: Opt: Open pre-trained transformer
  language models.
\newblock In: Proceedings of the Conference on Empirical Methods in Natural
  Language Processinng (2022)

\bibitem{zhang2022ow}
Zhang, Y., Zhang, X.Y., Shi, H.: Ow-tal: Learning unknown human activities for
  open-world temporal action localization.
\newblock Pattern Recognition p. 109027 (2022)

\bibitem{zhao2019hacs}
Zhao, H., Torralba, A., Torresani, L., Yan, Z.: Hacs: Human action clips and
  segments dataset for recognition and temporal localization.
\newblock In: Proceedings of the International Conference on Computer Vision,
  pp. 8668--8678 (2019)

\bibitem{zhao2020bottom}
Zhao, P., Xie, L., Ju, C., Zhang, Y., Wang, Y., Tian, Q.: Bottom-up temporal
  action localization with mutual regularization.
\newblock In: Proceedings of the European Conference on Computer Vision, pp.
  539--555. Springer (2020)

\bibitem{zhao2021soda}
Zhao, T., Han, J., Yang, L., Wang, B., Zhang, D.: Soda: Weakly supervised
  temporal action localization based on astute background response and
  self-distillation learning.
\newblock International Journal of Computer Vision \textbf{129}(8), 2474--2498
  (2021)

\bibitem{zhao2017temporal}
Zhao, Y., Xiong, Y., Wang, L., Wu, Z., Tang, X., Lin, D.: Temporal action
  detection with structured segment networks.
\newblock In: Proceedings of the International Conference on Computer Vision,
  pp. 2914--2923 (2017)

\bibitem{zheng2020distance}
Zheng, Z., Wang, P., Liu, W., Li, J., Ye, R., Ren, D.: Distance-iou loss:
  Faster and better learning for bounding box regression.
\newblock In: Proceedings of the AAAI Conference on Artificial Intelligence,
  pp. 12993--13000 (2020)

\bibitem{zhou2019learn}
Zhou, K., Yang, J., Loy, C.C., Liu, Z.: Learning to prompt for vision-language
  models.
\newblock In: International Journal of Computer Vision, pp. 2337--2348 (2019)

\bibitem{zhou2022conditional}
Zhou, K., Yang, J., Loy, C.C., Liu, Z.: Conditional prompt learning for
  vision-language models.
\newblock In: Proceedings of the IEEE Conference on Computer Vision and Pattern
  Recognition, pp. 16816--16825 (2022)

\bibitem{zhu2022prompt}
Zhu, B., Niu, Y., Han, Y., Wu, Y., Zhang, H.: Prompt-aligned gradient for
  prompt tuning.
\newblock arXiv preprint arXiv:2205.14865  (2022)

\bibitem{zhu2016learning}
Zhu, X., Loy, C.C., Gong, S.: Learning from multiple sources for video
  summarisation.
\newblock International Journal of Computer Vision \textbf{117}, 247--268
  (2016)

\bibitem{zhu2021enriching}
Zhu, Z., Tang, W., Wang, L., Zheng, N., Hua, G.: Enriching local and global
  contexts for temporal action localization.
\newblock In: Proceedings of the International Conference on Computer Vision,
  pp. 13516--13525 (2021)

\end{thebibliography}

\end{document}